\documentclass[journal]{IEEEtran}
\usepackage{cite}

\ifCLASSINFOpdf
  \usepackage[pdftex]{graphicx}
  \graphicspath{{figures/}}
\else
\fi
\usepackage{amsmath}
\usepackage{amssymb}
\usepackage{algorithm}
\usepackage{algpseudocode}

\algnewcommand\algorithmicforeach{\textbf{for each}}
\algdef{S}[FOR]{ForEach}[1]{\algorithmicforeach\ #1\ \algorithmicdo}
\algdef{SE}[DOWHILE]{Do}{doWhile}{\algorithmicdo}[1]{\algorithmicwhile\ #1}%

\algrenewcommand\algorithmicindent{1.0em}

\ifCLASSOPTIONcompsoc
  \usepackage[caption=false,font=normalsize,labelfont=sf,textfont=sf]{subfig}
\else
  \usepackage[caption=false,font=footnotesize]{subfig}
\fi
\usepackage{booktabs,makecell}
\usepackage{siunitx}

\usepackage{multirow}
\usepackage{hhline}
\usepackage[table,xcdraw]{xcolor}

\usepackage{url}

\makeatletter
\let\NAT@parse\undefined
\makeatother
\usepackage{hyperref}  

\usepackage{array}
\newcolumntype{L}[1]{>{\raggedright\arraybackslash}p{#1}}
\newcolumntype{C}[1]{>{\centering\arraybackslash}p{#1}}
\newcolumntype{R}[1]{>{\raggedleft\arraybackslash}p{#1}}

\usepackage{rotating,siunitx}

\begin{document}
%
\title{Parallel Structure from Motion for UAV Images via Weighted Connected Dominating Set}
%
%
%

\author{San~Jiang,
	Qingquan~Li,
	Wanshou~Jiang,
	Wu~Chen
	\thanks{S. Jiang and W. Chen are with the Department of Land Surveying and Geo-Informatics, The Hong Kong Polytechnic University, Hong Kong 999077, China; S. Jiang is also with the School of Computer Science, China University of Geosciences, Wuhan 430074, China. E-mail: \textit{jiangsan}@cug.edu.cn, \textit{wu.chen}@polyu.edu.hk.}
	\thanks{Q. Li is with the College of Civil and Transportation Engineering, Shenzhen University, Shenzhen 518060, China, and also with the Guangdong Laboratory of Artificial Intelligence and Digital Economy (Shenzhen), Shenzhen 518060, China. E-mail: \textit{liqq}@szu.edu.cn.}
	\thanks{W. Jiang is with the State Key Laboratory of Information Engineering in Surveying, Mapping and Remote Sensing, Wuhan University, Wuhan 430072, China. E-mail: \textit{jws}@whu.edu.cn.}
	\thanks{\textit{Corresponding author: Wu Chen}}}

%
%

\markboth{Journal of \LaTeX\ Class Files}%
{Shell \MakeLowercase{\textit{et al.}}: Bare Demo of IEEEtran.cls for IEEE Journals}
%



\maketitle

\begin{abstract}
Incremental Structure from Motion (ISfM) has been widely used for UAV image orientation. Its efficiency, however, decreases dramatically due to the sequential constraint. Although the divide-and-conquer strategy has been utilized for efficiency improvement, cluster merging becomes difficult or depends on seriously designed overlap structures. This paper proposes an algorithm to extract the global model for cluster merging and designs a parallel SfM solution to achieve efficient and accurate UAV image orientation. First, based on vocabulary tree retrieval, match pairs are selected to construct an undirected weighted match graph, whose edge weights are calculated by considering both the number and distribution of feature matches. Second, an algorithm, termed weighted connected dominating set (WCDS), is designed to achieve the simplification of the match graph and build the global model, which incorporates the edge weight in the graph node selection and enables the successful reconstruction of the global model. Third, the match graph is simultaneously divided into compact and non-overlapped clusters. After the parallel reconstruction, cluster merging is conducted with the aid of common 3D points between the global and cluster models. Finally, by using three UAV datasets that are captured by classical oblique and recent optimized views photogrammetry, the validation of the proposed solution is verified through comprehensive analysis and comparison. The experimental results demonstrate that the proposed parallel SfM can achieve 17.4 times efficiency improvement and comparative orientation accuracy. In absolute BA, the geo-referencing accuracy is approximately 2.0 and 3.0 times the GSD (Ground Sampling Distance) value in the horizontal and vertical directions, respectively. For parallel SfM, the proposed solution is a more reliable alternative.
\end{abstract}


%
\IEEEpeerreviewmaketitle

\section{Introduction}
\label{sec:1}

\IEEEPARstart{U}{AV} (Unmanned aerial vehicle) has become one of the widely used remote sensing (RS) platforms in the field of photogrammetry and remote sensing. Compared with satellite and aerial RS platforms, UAVs have the characteristics of high flexibility, high timeliness, and high resolution \cite{jiang2021unmanned}, which have been used in transmission line inspection \cite{jiang2017uav,huang2020model}, precision agriculture management \cite{zheng2021growing}, and cultural heritage document \cite{bakirman2020implementation}. Efficient and accurate image orientation plays a critical role to ensure their applications in different domains.

Nowadays, SfM (Structure from Motion) has become the core technique for UAV image orientation \cite{jiang2018efficient}. Different from traditional POS (Positioning and Orientation System) aided aerial triangulation (AT), SfM can resume camera poses and scene 3D points from overlapped and unordered images without good initial values of unknown parameters \cite{snavely2006photo}, which has been implemented in well-known software packages as the AT module. According to the strategy for parameter initialization, SfM can be divided into three major groups, i.e., incremental SfM, global SfM, and hybrid SfM \cite{cui2017hsfm}. Compared with the other methods, incremental SfM has the advantages of robustness to outliers and high orientation precision, which is achieved through increasingly registering images for parameter initialization and iteratively executing bundle adjustment (BA) for parameter optimization \cite{jiang2020effcient}. This strategy, however, sacrifices the efficiency of image orientation. With the increase in data volumes and spatial resolutions, the capability for efficient and accurate orientation of UAV images becomes more and more important for modern SfM systems.

In the literature, different techniques have been designed to address the issues in the SfM-based image orientation pipeline. On the one hand, some research focused on the acceleration of BA optimization, including GPU (Graphics Processing Unit) and CPU (Central Processing Unit) based hardware acceleration \cite{wu2011multicore} and the simplification \cite{sun2016rba} and optimization \cite{lourakis2009sba} of the BA mathematical model. However, these solutions cannot address the inherent drawback caused by the sequential constraint of incremental SfM. On the other hand, the divide-and-conquer strategy has been extensively adopted to break down the sequential constraint of incremental SfM \cite{bhowmick2014divide}. The core idea is to divide the large-scale orientation problem into small-size and easy-to-process sub-problems, and the entire model is created by merging sub-models that can be reconstructed in an efficient distributed or parallel mode. Sub-model merging becomes the main difficulty. Simple methods utilize common camera poses or scene 3D points between sub-models, and the hierarchical methods organize sub-models as a tree structure and achieve model merging from bottom leaf nodes to the top root node. Their performance depends on the overlap structures between sub-models. For reliable model merging, global model constrained methods have also been documented, e.g., MLST (maximal leaf spanning tree) \cite{snavely2008skeletal} and MCSD (minimal connected dominating set) \cite{havlena2010efficient}. These methods, however, are designed for landmark images with extremely high redundancy and overlap degrees.

Considering the parallelism characteristic of the divide-and-conquer strategy, this study proposes a parallel incremental SfM solution for UAV images. The core idea of the proposed solution is to extract a global model from the image topological connection network (TCN), which is utilized as the global geometric constraint to assist model merging. Our main contributions are summarized as follows: (1) We propose a weighted connected dominating set (WCDS) algorithm to extract the global model, in which both node redundancy and edge weight are considered for TCN vertex selection. The proposed algorithm can enhance the edge connection of the global model and increase the completeness of SfM reconstruction. (2) We implement a workflow to achieve parallel SfM reconstruction, in which the image TCN graph can be established efficiently through the vocabulary tree-based image retrieval technique, and common 3D points between sub-models can be found efficiently through on-demand correspondence graph generation. (3) We verify the validation and demonstrate the performance of the proposed parallel SfM solution by using UAV datasets that are captured by both classical oblique and recent optimized views photogrammetry.

This paper is organized as follows. Section \ref{sec:2} gives a literature review that relates to SfM efficiency acceleration. Section \ref{sec:3} presents the workflow and detailed procedure of the proposed parallel SfM solution. By using real UAV datasets, a comprehensive analysis and comparison are presented in Section \ref{sec:4}. Finally, Section \ref{sec:5} presents the conclusions of this study and improment plans for future study.

\section{Related work}
\label{sec:2}

This study focuses on the efficiency improvement of incremental SfM. In the literature, existing solutions can be categorized into two major groups, i.e., BA acceleration methods, and divide-and-conquer methods. Thus, the literature review is conducted from these two aspects as presented in the following subsections.

\subsection{BA acceleration methods}
\label{sec:2.1}

Incremental SfM depends on the iterative local and global BA to increase the precision of newly registered images and decrease the accumulated error of the final model. Iterative BA is a time-consumption step that dramatically degenerates the performance of image orientation. Therefore, BA acceleration is the most direct solution for SfM acceleration, including hardware acceleration, BA model simplification, and BA model optimization.

\textbf{Hardware acceleration}. Hardware techniques have tremendous development in recent years, which have been exploited to improve the efficiency of bundle adjustment \cite{choudhary2012visibility,hansch2016modern,liu2012hybrid,wu2011multicore,zheng2017new}. In the work of PBA (parallel bundle adjustment), \cite{wu2011multicore} proposed using both multicore CPU and multicore GPU to solve the BA problem with high parallelism capability, in which the matrix-vector production operation is restructured to adapt to the hardware parallelizable mechanism. In the work of \cite{zheng2017new}, the preconditioned conjugate gradient (PCG) and GPU-based parallel computing techniques were utilized to implement an efficient BA algorithm that was used for efficient orientation for UAV images. In conclusion, orders of speedup ratios can be achieved via recent high-performance computing systems.

\textbf{BA model simplification}. In contrast to hardware acceleration, BA model simplification is another commonly used strategy to decrease the computational complexity of BA problems. Existing solutions are usually achieved by either decreasing the number of camera parameters \cite{rupnik2013automatic,sun2016rba} or the number of 3D point structures \cite{cefalu2016structureless,rodriguez2011reduced}. In the work of \cite{sun2016rba}, a simplified BA model, termed RBA (reduced bundle adjustment), was proposed to process multi-camera oblique photogrammetric images, in which the pose of oblique cameras was simplified as the constant relative poses between oblique and vertical images and the absolute pose of vertical cameras. RBA reduces the total number of camera parameters in the BA optimization problem. Other solutions attempt to reduce the number of 3D points involved in the BA optimization since the parameters of 3D points are extremely larger than that of camera poses. This strategy is implemented by selecting the most useful tracks for image orientation \cite{cao2019fast,zhang2016efficient} or merely optimizing camera parameters in the BA problem, which is termed the structure-less SfM technique \cite{cefalu2016structureless}.

\textbf{BA model optimization}. For image orientation, the BA optimization is usually modeled as a joint minimization of reprojection errors between predicted and observed points, and it is solved as a nonlinear least-square problem. The sparse property of the normal equation in BA optimization can be utilized to decrease the number of involved parameters. On the one hand, the sparse property between cameras and 3D points is exploited by consecutive solving camera poses and 3D points, such as the SBA (sparse bundle adjustment) software package \cite{lourakis2009sba}. On the other hand, the sparse connection between cameras has been further used to handle the block data of SBA efficiently, which was released in the sSBA (sparse SBA), and it outperforms SBA by an order of magnitude. Except for these two packages, g2o (general graph optimization) is another well-known general BA optimization package that decreases the computational complexity of traditional BA solvers by exploiting the structure features of the BA problem \cite{kummerle2011g2o}.

\subsection{Divide-and-conquer methods}
\label{sec:2.2}

BA acceleration cannot break down the sequential constraint of incremental SfM, and they depend on high hardware requirements for increasingly large-scale image orientation. Thus, divide-and-conquer methods have been proposed, which mainly consist of simple methods, hierarchical methods, and global model constrained methods. For these methods, pair-wise matches are first structured as an undirected weighted match graph, in which vertices indicate images, and edges weighted by matches are added for matched image pairs.

\textbf{Simple methods}. In this category, the initial match graph is first divided into sub-clusters with compact edge connections and small image size, and cluster merging is then achieved by using common camera poses or 3D points between sub-models \cite{bhowmick2014divide,chen2020graph,lu2019block,xu2021robust,zhu2017parallel}. In the work of \cite{bhowmick2014divide}, the initial match graph is cut into small clusters through the normalized-cut (NC) algorithm \cite{shi2000normalized}, in which edge weights are assigned as the similarity score of vocabulary tree based image retrieval. After the image orientation of each component, the entire scene is created by merging sub-models through their epipolar relationships. In the work of \cite{zhu2017parallel}, two constraints, termed size constraint and completeness constraint, are used to divide the match graph into clusters with desired image number inside each cluster and enough image overlap between two clusters. Instead of the NC algorithm for scene clustering, \cite{lu2019block} proposed using the matrix band reduction (MBR) algorithm because it can generate clusters with equal size and compact structure. Considering that the order of cluster merging affects the completeness and accuracy of the final model, \cite{chen2020graph} proposed the DagSfM algorithm that incorporates image graph and cluster graph into the steps of scene clustering and cluster merging, respectively. To improve the robustness of scene clustering and merging, \cite{xu2021robust} developed an automatic and dynamic strategy to split match graphs and merge sub-models, as well as evaluation metrics to search unreliable models.

\textbf{Hierarchical methods}. Instead of simply dividing the match graph into clusters at one time, hierarchical methods adopt a tree structure to organize scene clusters from bottom leaf nodes to the top root node, and cluster merging is executed by sewing up nodes from lower levels to higher levels. Hierarchical methods have been used as early as the arising of the SfM technique \cite{shum1999efficient}. In the latter work of \cite{farenzena2009structure,gherardi2010improving,toldo2015hierarchical}, the hierarchical cluster strategy was proposed for computing structure and motion, in which image similarity is calculated by using both correspondence number and their spatial distribution. The proposed SfM solution has been implemented in the commercial software package, termed SAMANTHA . Similarly, \cite{ni2012hypersfm} adopted the divide-and-conquer manner to recursively divide an SfM problem into sub-maps. Compared with simple methods, hierarchical methods avoid the difficulties of selecting rational cluster size and determining the merging sequence.

\textbf{Global model constrained methods}. Hierarchical methods inherently lack the global view of the reconstruction problem since they only iteratively merge sub-models from bottom nodes to the top root node. In the literature, some other research proposes merging clusters from the middle tree level or using the global model to restrict the clustering merging. In the work of \cite{bhowmick2017divide}, base clusters were defined as the middle tree nodes with a pre-defined number of images, and the subsequent cluster merging started from these base clusters instead of the classical methods that start from tree leaf nodes. Inspired by the preemptive matching  in \cite{wu2013towards}, \cite{shah2014multistage} first reconstructed a coarse and global model by using features with large scales, and the orientation of the remaining images was achieved in a parallel way through the direct resection using existing 3D-2D correspondences. In the work of \cite{havlena2010efficient}, a graph optimization algorithm based on MCDS (minimal connected dominating set) was designed to reduce the number of vertices in the match graph and decrease the complexity of the subsequent SfM reconstruction. The proposed solution was verified by using landmark images with high redundancy from internet communities.

\section{Parallel structure from motion}
\label{sec:3}

\begin{figure}[t!]
	\centering
	\includegraphics[width=0.5\textwidth]{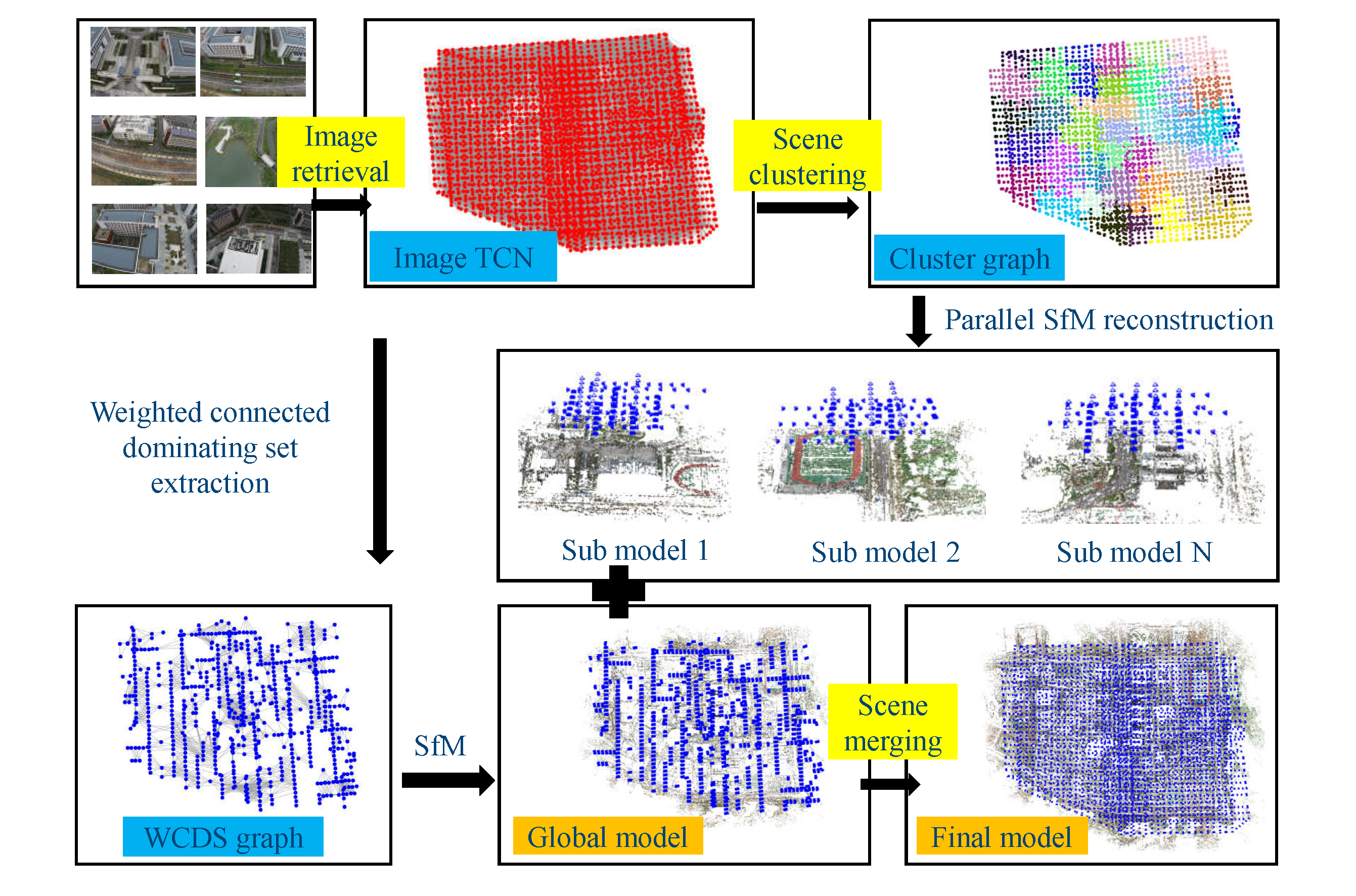}
	\caption{The workflow of the proposed parallel SfM solution.}
	\label{fig:figure1}
\end{figure}

Figure \ref{fig:figure1} presents the workflow of the proposed parallel Structure from Motion solution. The inputs of the parallel SfM solution are only UAV images and a pre-trained vocabulary tree without any other auxiliary data. The workflow consists of three major steps. First, overlapped match pairs are selected through vocabulary tree-based image retrieval, and an image TCN is then constructed as an undirected weighted graph, which establishes image connection and plays as the basic structure for global model extraction and scene clustering. Second, an algorithm, termed WCDS, is proposed to extract the global graph from the image TCN, which would be reconstructed and used as the global geometric constraint to guide scene merging. Meanwhile, the image TCN is simultaneously divided into small and compact clusters that can be reconstructed efficiently and accurately. Third, by using the WCDS graph and divided clusters, parallel SfM reconstruction is executed to generate the global model and cluster models, which are merged to generate the final model. This section first describes the used incremental SfM engine and then presents the workflow of the proposed parallel SfM solution.

\subsection{Incremental structure from motion}
\label{sec:3.1}

The workflow of the used incremental Structure from Motion engine consists of two modules, i.e., correspondence matching and incremental reconstruction \cite{jiang2020effcient}. The former aims to search for accurate and enough correspondences between overlapped match pairs, which can recover the relative geometry of two images; the latter is used to resume and refine camera poses and scene 3D points by using pair-wise geometry and BA optimization. In the field of photogrammetry and remote sensing, local feature-based image matching has become the widely used technique for correspondence matching because of their invariance to image rotation, scale difference, and even the changes of view-points and illuminations. In this study, the Root-SIFT with the L1-norm distance metric \cite{dong2015domain} has been adopted for feature extraction and matching.


After correspondence matching, incremental reconstruction is executed to recover camera poses and 3D points. Similar to BA optimization in photogrammetry, pair-wise feature matches are first tied to generate tracks, which corresponds to a set of matched feature points that see the same 3D location \cite{zhang2016efficient}. Incremental SfM is then started by selecting two seed images that have a large enough intersection angle and a sufficient number of well-distributed matches, and the global model is constructed by resuming their relative poses and 3D points. For unregistered images, an iterative procedure is executed by searching for the next-best image that observes the largest number of resumed 3D points and registering the image into the reconstructed model. Meanwhile, both local and global BA optimization are alternately executed to refine the poses of newly added images and decrease accumulated errors \cite{snavely2008skeletal}.

For local and global BA optimization, the problem for refining camera poses and scene 3D points is formulated as a joint minimization of the reprojection function \cite{triggs1999bundle}, where the sum of errors between the track projections and their corresponding image points is minimized. The object function of BA optimization is presented by Equation \ref{eq:1}

\begin{equation}
	\min\limits_{C_j,X_i}\sum_{i=1}^n\sum_{j=1}^m\rho_{ij}\parallel P(C_j,X_i)-x_{ij} \parallel ^2
	\label{eq:1}
\end{equation}
where $X_i$ and $C_j$ indicate a 3D point and a camera, respectively; $P(C_j, X_i)$ is the projection of point $X_i$ on camera $C_j$; $x_{ij}$ is an observed image point; $\left\|\bullet\right\|$ denotes L2-norm; $\rho_{ij}$ is an indicator function with $\rho_{ij}=1$ if point $X_i$ is visible in camera $C_j$; otherwise $\rho_{ij}=0$. In this study, the Ceres Solver package is used for BA optimization.

\subsection{Match graph construction via vocabulary tree-based image retrieval}
\label{sec:3.2}

Match graph is used as the basic structure for clustering and merging in the parallel SfM solution, as well as guiding feature matching for efficiency improvement \cite{jiang2017Efficient}. Before match graph construction, match pair selection should be conducted by using an efficient strategy. For most photogrammetric systems, onboard POS data is widely used to predict match pairs, in which image footprints on an average elevation plane are generated through the spatial intersection, and overlapped match pairs are determined through the intersection test between image footprints \cite{jiang2017board}. Although high efficiency can be achieved, these methods, however, depend on the precision of onboard POS data and the elevation of test sites, and they cannot adapt to data acquisition in optimized views photogrammetry.

In this study, vocabulary tree-based image retrieval has been utilized to perform match pair selection. In contrast to other methods, vocabulary tree-based image retrieval does not rely on any other auxiliary data but the images themselves. The core idea of vocabulary tree-based image retrieval is to represent each image as a BoW (Bag-of-Words) vector, and the problem of finding match pairs is cast as searching images with the closest BoW vectors between the query and database images \cite{jiang2022leveraging,nister2006scalable}. For match pair selection, a pre-trained and open-released vocabulary tree has been utilized in this study, which has 256 thousand visual words that are trained using images from internet communities. To achieve the highest efficiency, the number of used features for image indexing is set as 1500, and the number of retrieved images is configured as 100, which recovers enough match pairs.

After match pair selection, feature matching is then executed by using the SIFTGPU library for efficiency improvement, which is finally used for match graph construction. Suppose that $I=\{i_i\}$ and $P=\{p_{ij}\}$  are the images of size $n$ and match pairs of size $m$, respectively; the match graph is represented by an undirected graph $G=(V,E)$, where $V$ and $E$ stand for the vertex set and edge set of the undirected graph $G$, respectively. Thus, the match graph is constructed as follows: define a vertex $v_i$ for each image $i_i$, such that $V=\{v_i,i=1,2,...,n\}$; an undirected edge $e_{ij}$ that connects vertices $v_i$ and $v_j$ is added for each of the $m$ match pairs $p_{ij}$, such that $E=\{e_{ij},i,j=1,2,...,n,i \neq j\}$. In addition, a weight value $w_{ij}$ is assigned to the edge $e_{ij}$.

In the context of SfM reconstruction, the weight value $w_{ij}$ encodes the connection strength of the corresponding edge $e_{ij}$. As feature matching has been conducted, it is rational to calculate the edge weight by using the number of feature matches. For image orientation, both the number and distribution of feature matches affect the robustness and precision of two view geometry estimation. Thus, the weight value $w_{ij}$ is calculated according to Equation \ref{eq:2}

\begin{equation}
	w_{ij}=R_{ew}*w_{inlier}+(1-R_{ew})*w_{overlap}
	\label{eq:2}
\end{equation}
where $w_{inlier}$ and $w_{overlap}$ are respectively weight items calculated by the number of feature matches and the overlap ratio between the convex of feature matches and the area of image planes; $R_{ew}$ is the weight ratio of these two items, which is set as 0.5 in this study. Thus, the weight value $w_{ij}$ is a linear combination of these two items, which are calculated according to Equations \ref{eq:3} and \ref{eq:4}

\begin{equation}
	w_{inlier}=\frac{log(N_{inlier})}{log(N_{max\_inlier})}
	\label{eq:3}
\end{equation}
\begin{equation}
	w_{overlap}=\frac{CH_i+CH_j}{A_i+A_j}
	\label{eq:4}
\end{equation}
where $N_{inlier}$ and $N_{max\_inlier}$ respectively indicate the number of feature matches of the current match pair $p_{ij}$ and the maximum number of feature matches of all match pairs $P$; $CH_i$ and $CH_j$ are the convex hull areas of feature matches between images $i_i$ and $i_j$, respectively; $A_i$ and $A_j$ are the areas of image planes of images $i_i$ and $i_j$, respectively. In this study, the convex hull of matched feature points is detected by using the Graham-Andrew algorithm \cite{andrew1979another}. Noticeably, before the construction of the match graph, the match pairs with the number of matched features less than 50 are removed due to two main reasons. On the one hand, false matches may exist in these match pairs; on the other hand, they have a weaker connection in the subsequent global model construction.


\subsection{Parallel structure from motion}
\label{sec:3.3}

Parallel structure from motion can be implemented by dividing the original problem into some small-size and compact clusters that could be independently reconstructed based on the incremental SfM engine. In this study, both scene clustering and cluster merging are achieved by using the match graph. First, a weighted connected dominating set is extracted from the initial match graph, which consists of a well connected subset of vertices and is used to reconstruct a global model to assist cluster merging. Second, compact clusters are generated by splitting the match graph into some independent parts without overlap vertices, which can be reconstructed efficiently under the memory limitation of computers. Finally, scene merging is conducted to generate the final model using the common 3D points between the global model and cluster models. The details of each step are described in the following subsections.

\subsubsection{Weighted connected dominating set for the global model}
\label{sec:3.3.1}

Cluster merging is a non-trivial task in the SfM based on the divide-and-conquer strategy. In this study, the core idea of the proposed solution is to create a global model over the entire scene, and subsequent cluster merging can be constrained with the global model. To construct the global model, a tradeoff between two contradictory conditions should be made. On the one hand, the global model should be created by using as least number of images as possible as that the lowest computational costs can be consumed; on the other hand, the images involved in the global model should have strong and enough connections to ensure successful SfM reconstruction, which in turn would increases the number of images.

\begin{figure}[t!]
	\centering
	\includegraphics[width=0.5\textwidth]{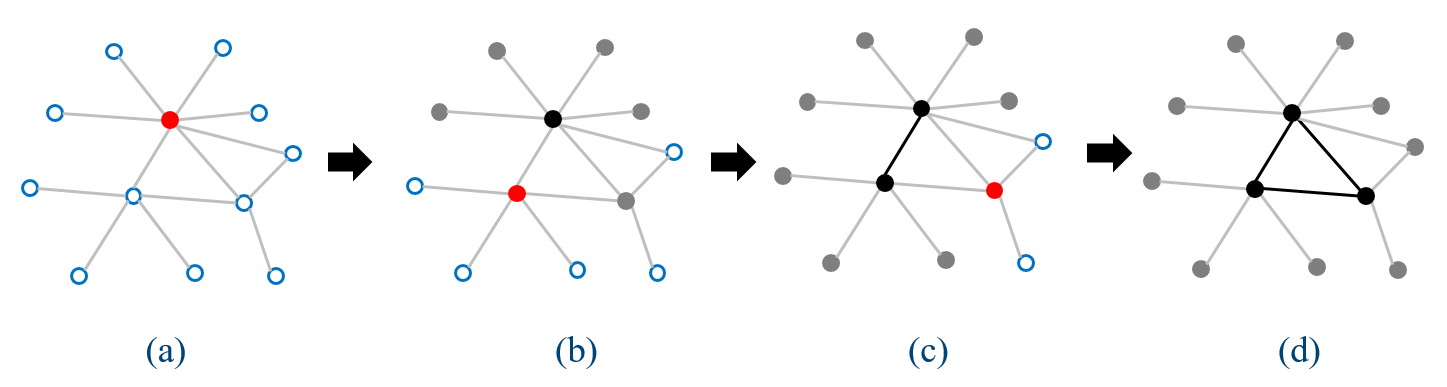}
	\caption{The illustration of extracting MCDS: (a) initial match graph; (b) – (d) three vertices are iteratively appended into MCDS, and black nodes indicate the vertices of the extracted MCDS.}
	\label{fig:figure4}
\end{figure}

For the first condition, the purpose of extracting a minimal subset of images is the same as the minimal connected dominating set (MCDS) algorithm \cite{guha1998approximation}. Suppose that the match graph is represented by $G=(V,E)$. MCDS aims to find a minimal subset of vertices $V_m$ from $V$ in graph $G$, such that the vertex set $V_m$ meets two conditions: first, the graph $G_m$ deduced from the vertex set $V_m$ is connected, which means that all vertices in $V_m$ are tied together; second, for any vertex $v_i$ in $V$, it belongs to the vertex set $V_m$; otherwise, it is the neighbor of at least one vertex in the vertex set $V_m$. Figure \ref{fig:figure4} illustrates the procedure of extracting MCDS from the match graph, which consists of four major steps as follows:

\begin{enumerate}
	\item Initialization. In the beginning, all vertices in $V$ are labeled in while color, which indicates that all vertices are not visited. Meanwhile, set the vertex that has the largest number of white neighbors as the current vertex $v^*$, which is rendered in red color in Figure \ref{fig:figure4}(a).
	\item Scan and mark vertices. For the current vertex $v^*$, scan it (label it in black color). At the same time, mark the neighbors of $v^*$ (label them in gray color), as shown in Figure \ref{fig:figure4}(b).
	\item Find the next-current vertex. Among all marked vertices (vertices in gray color), set the vertex that has the largest number of white neighbors as the next-current vertex $v^*$, as shown in Figure \ref{fig:figure4}(b) and Figure \ref{fig:figure4}(c).
	\item Iteratively execution of steps (2) and (3) until no while vertices exist. The vertices in black color consist of the extracted MCDS, as shown in Figure \ref{fig:figure4}(d).
\end{enumerate}

In the MCDS algorithm, the purpose of extracting a minimal subset of vertices from the match graph is implemented by selecting the next-current vertex $v^*$ that has the largest number of white neighbors. This strategy, however, does not consider the second condition of creating the global model. That is, the images involved in the extracted vertex subset should have strong and enough connections since this condition ensures the precision and completeness of the final model in the context of SfM reconstruction. Thus, a weighted connected dominating set algorithm, termed WCDS, has been designed in this study. The core idea is to pose the edge weight constraint on the selection of the next-current vertex $v^*$. In particular, WCDS uses both the number of white neighbors and the edge weight between vertices to quantify the importance of gray vertices, as represented in Equation (5)

\begin{equation}
	w_{ver}=R_{vw}*w_{ngb}+(1-R_{vw})*w_{ij}
	\label{eq:5}
\end{equation}
where $w_{ngb}$ is the weight item calculated from the number of white neighbors; $w_{ij}$ is the edge weight that has been assigned to the corresponding edge $e_{ij}$ in match graph construction, as represented in Equation \ref{eq:2}; $R_{vw}$ is the weight ratio that controls the contribution of these two weight items, which ranges between 0.0 and 1.0. Noticeably, when $R_{vw}$ is set as 1.0, the classical MCDS is utilized; when $R_{vw}$ is set as 0.0, only edge weight affects the calculation of the gray vertex importance. In Equation \ref{eq:5}, the weight term $w_{ngb}$ has been normalized by Equation \ref{eq:6}, in which $N_{ngb}$ is the number of white neighbors for the current vertex, and $N_{max\_ngb}$ is the maximal number of white neighbors for all vertices in the initial match graph as shown in Figure \ref{fig:figure4}(a).

\begin{equation}
	w_{ngb}=\frac{N_{ngb}}{N_{max\_ngb}}
	\label{eq:6}
\end{equation}

Based on the weight calculation of gray vertices, the procedure of the proposed WCDS algorithm has been modified in step (3) when compared with the classical MCDS: for all marked vertices, calculate the gray vertex weight according to Equation \ref{eq:5}, and set the vertex with the highest importance value $w_{ver}$ as the next-current vertex $v^*$. When one marked gray vertex connects to more than one scanned black vertex as shown in Figure \ref{fig:figure4}(c), the vertex importance value of the marked gray vertex is calculated by using the largest edge weight. The main reason is that it is the strongest connection between the marked gray vertex and scanned black vertices.

\subsubsection{Scene clustering for parallel structure from motion}
\label{sec:3.3.2}

Scene clustering aims to divide the entire scene into some small-size and compact clusters that can be efficiently and accurately reconstructed. To achieve the highest efficiency, each image that is not registered in the global model can be seen as an individual cluster, whose camera poses can be resumed efficiently using the 3D-2D correspondences deduced from the reconstructed global model. However, considering that the global model only contains a very small fraction of images, it would be hard to search enough 3D-2D correspondences for unregistered images. Therefore, scene clustering is conducted to divide the match graph into some clusters, which can cover more 3D points in the global model.

Scene clustering has been implemented by splitting an initial match graph into some overlapped sub graphs under constraints, e.g., the size constraint and the completeness constraint. The former is used to control the size of each cluster for efficient parallel reconstruction; the latter is used to ensure enough common images between clusters for reliable model merging. Since the global model has been created to assist cluster merging, only the size constraint has been used in this study. For scene clustering, the normalized cut (NC) algorithm \cite{shi2000normalized} has been selected, which is prone to cut graph edges with smaller weights and generate compact clusters with strong inner connections. After scene clustering, each image is assigned to only one cluster, and each cluster can be reconstructed based on the parallel SfM engine.

\subsubsection{Scene merging with the global geometric constraint}
\label{sec:3.3.3}

After the global model and cluster models are reconstructed based on the incremental SfM engine, each reconstructed model has its own coordinate system. The reconstructed models should be merged into the same coordinate system to generate a complete model. In general, scene merging is achieved through common structures between two cluster models, e.g., camera poses and 3D points. Suppose that two models $m_s$ and $m_r$ are respectively defined as the source reconstruction and reference reconstruction; there are $n$ common structures between these two models, as represented by   $\{s_i\}$ and $\{r_i\}$, $i=1,2,...,n$; the transformation from $m_s$ to $m_r$ is formulated by a similarity transformation $T=[\lambda R|t]$. In general, $T$ can be computed by minimizing the objective function as shown by Equation (7)

\begin{equation}
	e^2(R,t,\lambda)=\frac{1}{n}\sum_{i=1}^{n}\parallel r_i-(\lambda Rs_i+t) \parallel ^2
	\label{eq:7}
\end{equation}
where $R$ is the rotation matrix; $t$ is the translation vector; $\lambda$ is the scaling factor; $e^2$ is the mean square error (MSE) of transforming structures $\{s_i\}$ to $\{r_i\}$ under transformation $T$, which can be robustly estimated using the SVD (singular value decomposition) algorithm \cite{umeyama1991least}. In this study, common 3D points between global and cluster reconstructions are used as the structure for similarity transformation estimation due to two main reasons. On the one hand, the number of 3D points is extremely larger than that of camera poses; on the other hand, 3D points are distributed more evenly in the global model.

Due to the existence of outliers in 3D points, the robust estimation technique based on the RANSAC framework has been utilized in similarity transformation estimation. Instead of using MSE for model verification, the bi-directional mean square reprojection error is utilized to calculate the transformation residuals under one hypothesis, as presented in Equation \ref{eq:8}. In this study, the maximum residual $r_i$ is set as 1.8 pixels to separate inliers from outliers.

\begin{equation}
\begin{split}
	r_i^2=\frac{1}{m+l}(\sum_{j=1}^{m} \parallel P(C_j,T*s_i)-x_{ij}\parallel ^2 \\
	+\sum_{j=1}^{l} \parallel P(C_j, T^{-1}*r_i)-x_{ij} \parallel ^2)
	\label{eq:8}
\end{split}
\end{equation}

\begin{figure}[t!]
	\centering
	\includegraphics[width=0.45\textwidth]{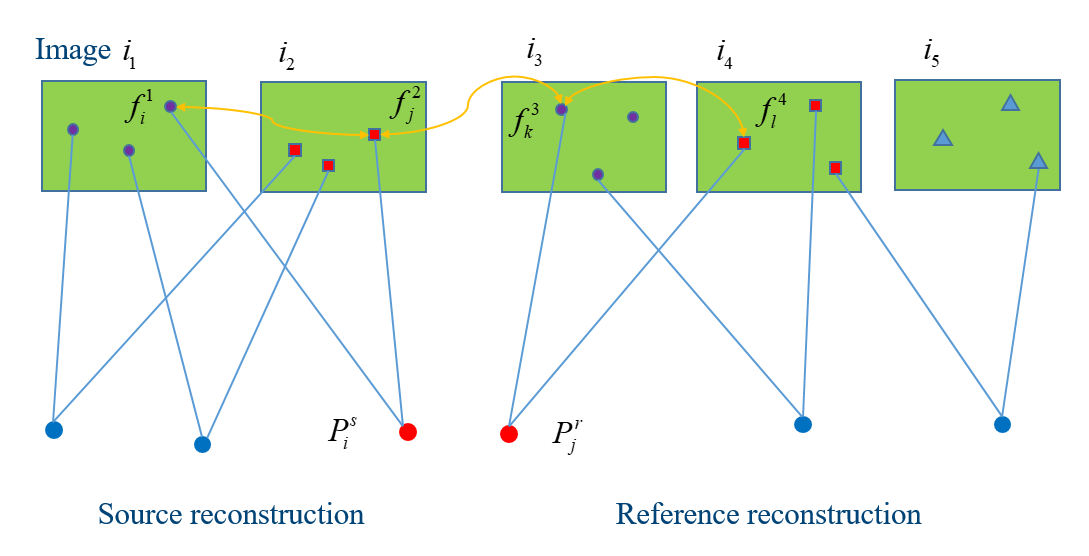}
	\caption{The illustration of searching common 3D points by creating the on-demand correspondence graph between source and reference reconstructions.}
	\label{fig:figure5}
\end{figure}

Common 3D points between the global and cluster reconstructions should be efficiently determined during scene merging. In general, common 3D points can be found by either using common images between reconstructions \cite{chen2020graph} or by merging tracks from global and cluster reconstructions \cite{zhu2017parallel}. These methods either require high overlap degrees or need large memory consumption. In this study, an on-demand correspondence graph is created to establish the mapping relationship between feature matches, as illustrated in Figure \ref{fig:figure5}. Two reconstructed models have respectively termed source and reference reconstructions. Within each reconstruction, the relationship between 2D feature points and 3D scene points has been established through the generated tracks in SfM reconstruction. In other words, the problem of finding common 3D points becomes establishing the mapping relationship between feature matches across these two reconstructions.

Considering that the number of images in the source reconstruction is much less than that in the reference reconstruction, the mapping relationship has been constructed by only using feature matches that are related to the images in the source reconstruction. As shown in Figure \ref{fig:figure5}, only images labeled $i_1$, $i_2$, $i_3$ and $i_4$ are used to construct the correspondence graph, and the mapping relationship between feature matches are indicated by the yellow lines. Using the established correspondence graph, common 3D points $C$ are found according to the procedure: (1) for each 3D point $P_i^s$ in the source reconstruction, find its related observations $O=\{f_i^1,f_j^2\}$; (2) for each observation in $O$, by using the mapping relationship in the correspondence graph, find its related feature matches $M=\{f_k^3\}$ from the images in the reference reconstruction; (3) for each feature match in  $M$, find its corresponding 3D points $P_j^r$ based on established tracks in the reference reconstruction; (4) add the common 3D point pair $(P_i^s,P_j^r)$ to $C$ if it does not exist in $C$. For scene merging, the proposed method based on on-demand correspondence graph has very high efficiency and low memory cost.

Two reconstructions can be merged by using their established similarity transformation within the robust RANSAC framework. Although all cluster reconstructions are transformed into the coordinate system of the base reconstruction, the merging sequence among the cluster reconstruction would affect the precision and completeness of the final model. Intuitively, the MSE from similarity transformation estimation may be a good metric to quantify the merging quality of two reconstructions, which has been used in other work \cite{chen2020graph}. This study, however, orders the merging sequence based on the number of common 3D points due to two main reasons. On the one hand, two reconstructions with more common 3D points are usually prone to be merged with high precision; on the other hand, the earlier more 3D points are merged into the global model, the more common 3D points can be found for the subsequent cluster models. After merging all cluster models, one final global BA is executed to optimize the camera poses and 3D points to refine the final model.

\subsection{Algorithm implementation}

Based on the principle of the proposed solution, this study has implemented the parallel SfM system for UAV images. We have used the C++ programming language to achieve the highest efficiency. For feature detection and matching, the used SIFTGPU has been accelerated by using CUDA (Compute Unified Device Architecture), and the default parameters are used as documented in \cite{wu2007siftgpu}. For SfM reconstruction, the open-source software package ColMap \cite{schonberger2016structure} has been used as the backend SfM engine. In this study, we have added the BROWN camera model that has been widely used in the compared commercial software packages and implemented an absolute BA optimizer to support geo-referencing accuracy assessment. In addition, camera intrinsic parameters have been calibrated and are fixed in BA optimization. The algorithmic implementation is presented in Algorithm 1

\begin{algorithm}[htp!]
	\caption{Parallel SfM based on WCDS}
	\label{alg:ParallelSfM}
	\hspace*{\algorithmicindent} \textbf{Input:} UAV images $I=\{i_i\}$; a pre-trained vocabulary tree $VocT$ \\
	\hspace*{\algorithmicindent} \textbf{Output:} reconstructed model $M$
	\begin{algorithmic}[1]
		\Procedure{TCNConstruction}{}
		\ForEach{$i_i \in I$}
		\State Extract SIFT features for image $i_i$
		\State Select the top-scale 1500 features of image $i_i$ and index them into $VocT$
		\EndFor
		\ForEach{$i_i \in I$}
		\State Query the top 100 similar images and add them to match pairs $P=\{p_{ij}\}$
		\EndFor
		\State Create $TCN$ graph using $P$ with edge weights calculated by Equation \ref{eq:2}
		\EndProcedure
	\end{algorithmic}
	\begin{algorithmic}[1]
		\Procedure{WCDSExtraction}{}
		\State Initialize all vertices in $TCN$ as white color
		\State Select the vertex with the most white neighbors as current vertex $v^*$
		\Do
		\State Scan vertex $v^*$ and mark its neighbors
		\State Update the weight of all gray vertices according to Equation \ref{eq:5}
		\State Set the gray vertex with the largest weight as next-current vertex $v^*$
		\doWhile{white vertex exists}
		\State Create the $WCDS$ graph using vertices in black color
		\EndProcedure
	\end{algorithmic}
	\begin{algorithmic}[1]
		\Procedure{ParallelReconstruction}{}
		\State Start a thread to reconstruct $WCDS$ based on incremental SfM
		\State Divide $TCN$ into non-overlapped clusters $C=\{c_j\}$ of size $m$
		\ForEach{$c_j \in C $}
		\State Start a thread to reconstruct $c_j$ based on incremental SfM
		\EndFor
		\EndProcedure
	\end{algorithmic}
	\begin{algorithmic}[1]
		\Procedure{ClusterMerging}{}
		\State Set $M=WCDS$
		\While{$C$ not empty}
		\State Set $P=\{\}$
		\ForEach{$c_j \in C $}
		\State Find common 3D points $p_j$ between $c_j$ and $M$, set $P=P+\{p_j\}$
		\EndFor
		\State Find $p_k^*$ that has the largest number of common 3D points
		\State Calculate transform $T_k$ using RANSAC according to Equation \ref{eq:8}
		\State Merge $c_k$ into $M=M+\{c_j\}$ and set $C=C-\{c_j\}$
		\EndWhile
		\State Conduct the final BA for the merge model $M$
		\EndProcedure
	\end{algorithmic}
\end{algorithm}

\section{Experimental results and discussions}
\label{sec:4}

In experiments, three UAV datasets are collected to evaluate the performance of the proposed parallel SfM solution. First, the influence of the weight ratio $Rvw$ on extracting the WCDS graph is analyzed in terms of the selected image ratio and connected image ratio. Second, memory consumptions are compared in the cluster merging with and without the on-demand correspondence graph strategy. Third, by using the optimal parameter setting, the SfM solution is executed to reconstruct the three UAV datasets. Finally, the proposed SfM solution is extensively compared with four open-source and commercial software packages, including ColMap \cite{schonberger2016structure}, DagSfM \cite{chen2020graph}, Agisoft Metashape \cite{2021Agisoft}, and Pix4dMapper \cite{2021Pix4Dmapper}, and the results are analyzed in relative BA without ground control points (GCPs) and absolute BA with GCPs.

\subsection{Test sites and datasets}
\label{sec:4.1}

Table \ref{tab:table1} presents the detailed information of these three UAV datasets. For outdoor data acquisition, multi-rotor UAVs are used in the three campaigns. The details of each dataset are described as follows:

\begin{table}[!b]
	\centering
	\caption{Detailed information of the three UAV datasets.}
	\label{tab:table1}
	\makebox[0.5\linewidth]{
		\begin{tabular}{l l l l}
			\toprule
			\textbf{Item Name} & \textbf{Dataset 1} & \textbf{Dataset 2} & \textbf{Dataset 3} \\
			\midrule
			UAV type & multi-rotor &  multi-rotor & multi-rotor \\
			Flight height (m) & 175 & 80 & - \\
			Camera mode & Sony NEX-7 & DJI FC6310R & DJI ZenmuseP1 \\
			Camera number & 5 & 1 & 1 \\
			Focal length (mm) & \begin{tabular}{@{}l@{}} nadir: 16 \\ oblique: 35 \end{tabular} & 24 & 35 \\
			Camera angle ($^\circ$) & \begin{tabular}{@{}l@{}} nadir: 0 \\ oblique: 45/$-$45 \end{tabular} & 35 & - \\
			Number of images & 750 & 3743 & 4030 \\
			Image size (pixel)  & 6000 $\times$ 4000 & 5472 $\times$ 3648 & 8192 $\times$ 5460 \\
			GSD (cm) & 4.3 & 2.6 & 1.2 \\
			\bottomrule
		\end{tabular}
	}
\end{table}

\begin{figure*}[ht!]
	\centering
	\subfloat[Dataset 1]{\includegraphics[height=0.25\textwidth]{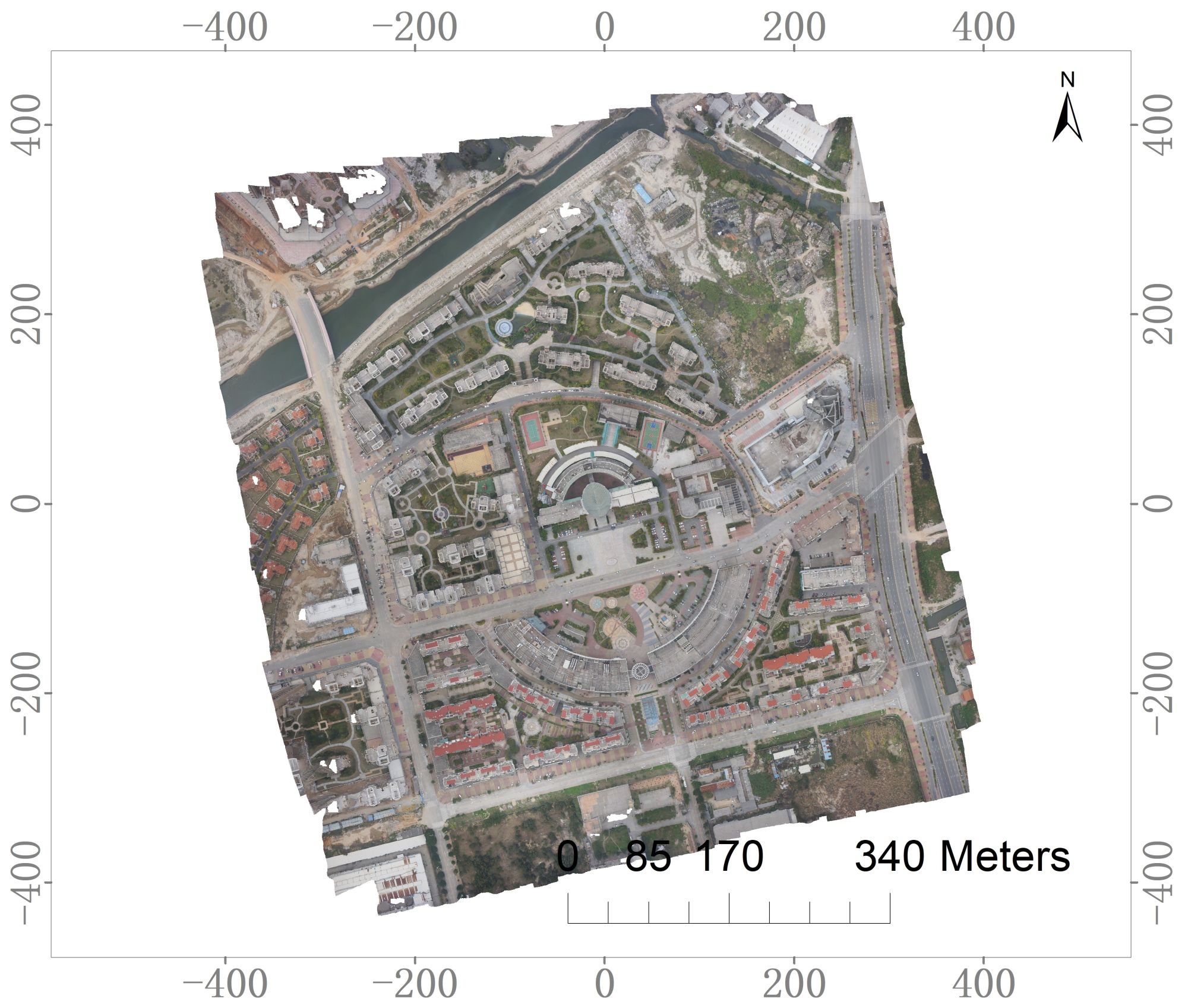}
		\label{fig:figure6-a}}
	\subfloat[Dataset 2]{\includegraphics[height=0.25\textwidth]{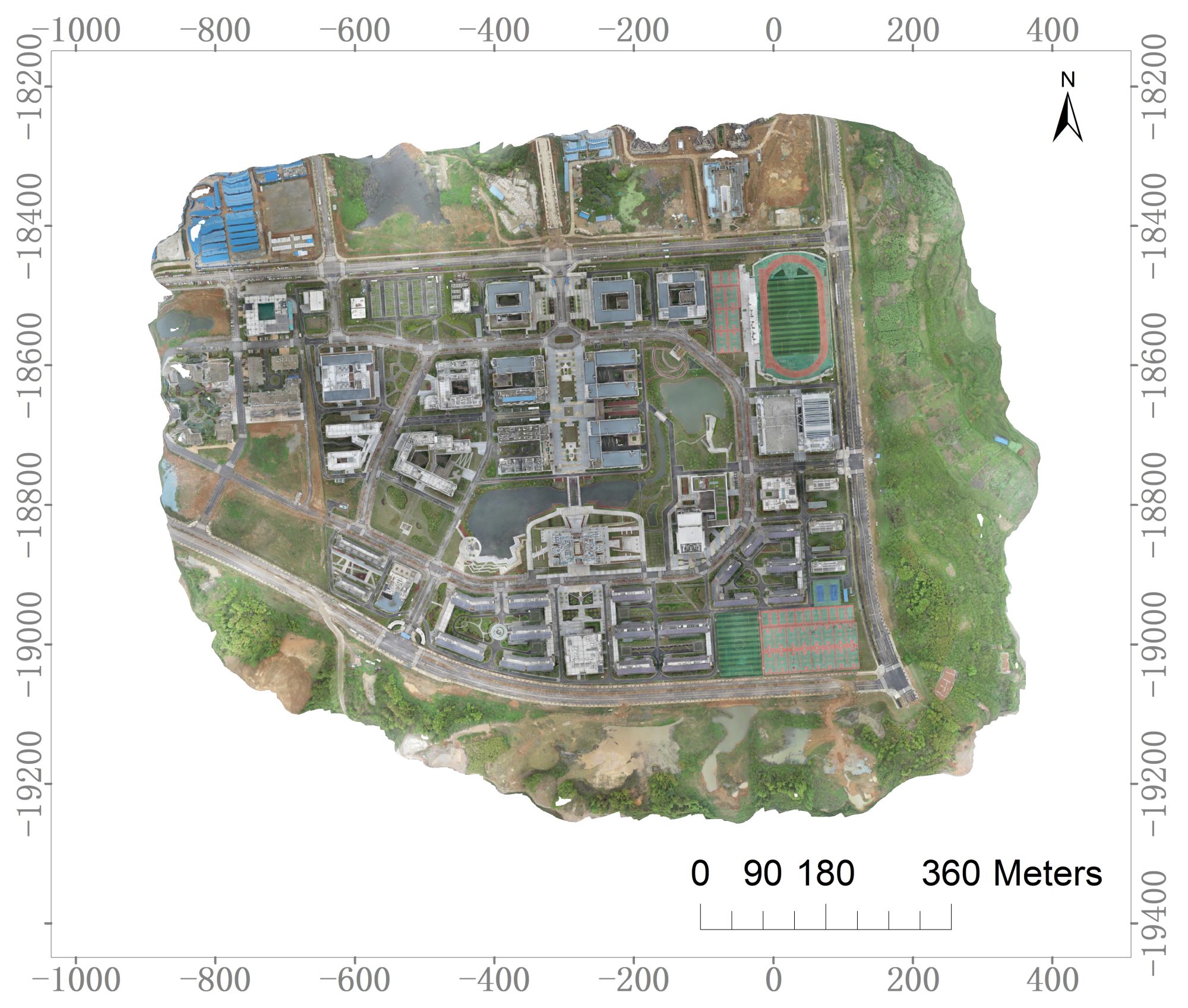}
		\label{fig:figure6-b}}
	\subfloat[Dataset 3]{\includegraphics[height=0.25\textwidth]{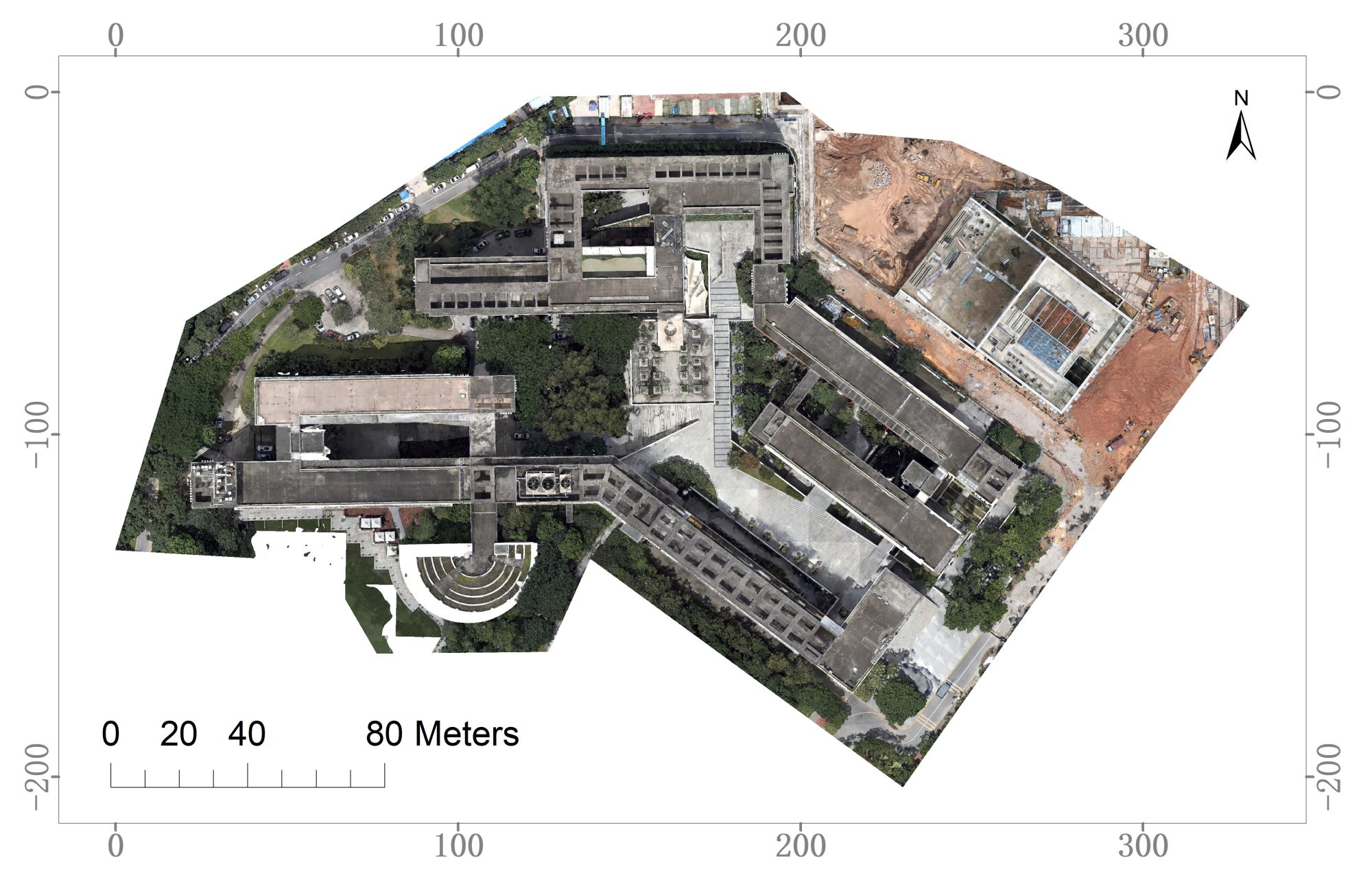}
		\label{fig:figure6-c}}
	\caption{The orthomosaics for the test sites of the three datasets.}
	\label{fig:figure6}
\end{figure*}

\begin{itemize}
	\item The first dataset covers an urban shopping plaza that is surrounded by high buildings, as shown in \ref{fig:figure6-a}. For image acquisition, a classical five-camera oblique photogrammetric system consisting of one nadir camera and four oblique cameras is utilized, where four oblique cameras are rotated 45° with respect to the nadir camera. The resolution of captured images is 6000 by 4000 pixels. Under the flight height of 175 m, a total number of 750 images are recorded with the GSD value of 4.3 cm.
	\item The second dataset covers a university campus, which includes dense low buildings, as shown in \ref{fig:figure6-b}. The surroundings of the campus are vegetation and bare land. By using a DJI Phantom 4 RTK UAV equipped with one DJI FC6310R camera, a total number of 3743 images with the dimension of 5472 by 3648 pixels are collected under the flight height of 80 m. The GSD of collected images is approximately 2.6 cm.
	\item The third dataset is recorded from a university campus, which is mainly covered by a complex building, as shown in \ref{fig:figure6-c}. In contrast to the other datasets, the optimized views photogrammetric technique \cite{zhou2020offsite} has been used for designing the UAV trajectory, which adjusts view points and directions according to the coarse geometric model of the test site. By using the DJI M300 RTK UAV equipped with one DJI Zenmuse P1 camera, a total number of 4030 images are recorded with the GSD of 1.2 cm under the view distance of approximately 80 m to ground targets. Besides, for the geo-reference accuracy assessment, 26 GCPs are surveyed using a total station, whose nominal horizontal and vertical accuracy is about 0.8 cm and 1.5 cm. GCPs are made by using artificial markers and distributed on ground, building facades and roofs.
\end{itemize}


\subsection{The influence of the weight ratio on the construction of the WCDS graph}
\label{sec:4.2}

In the construction of the WCDS graph, the weight ratio $R_{vw}$ controls the contribution of the number of white neighbors and the edge weight to quantify the importance of gray vertices. This section would analyze the influence of the weight ratio $R_{vw}$ on creating the WCDS graph.

\begin{figure}[!ht]
	\centering
	\subfloat[Selected image ratio]{\includegraphics[width=0.25\textwidth]{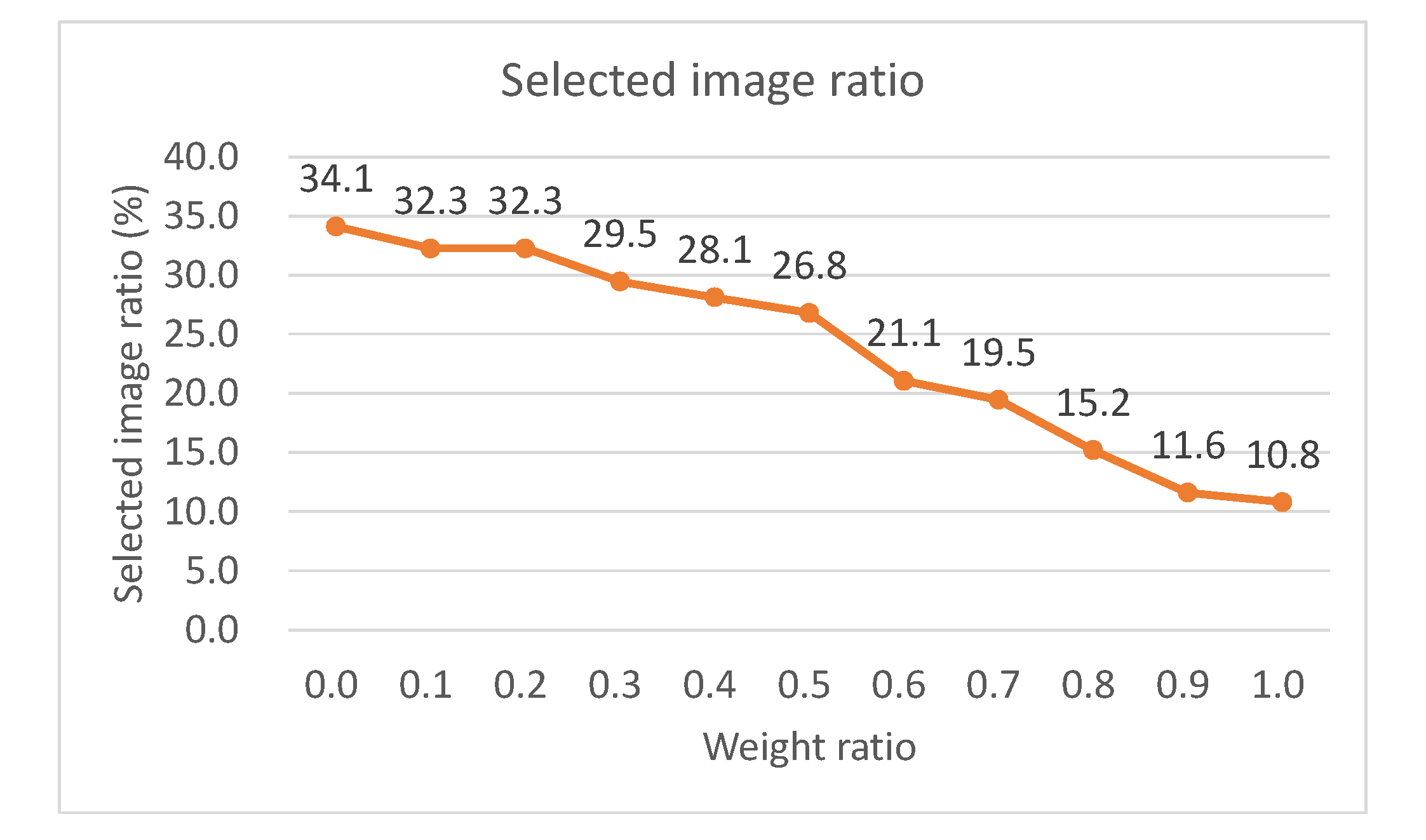}
		\label{fig:figure8-a}}
	\subfloat[Connected image ratio]{\includegraphics[width=0.25\textwidth]{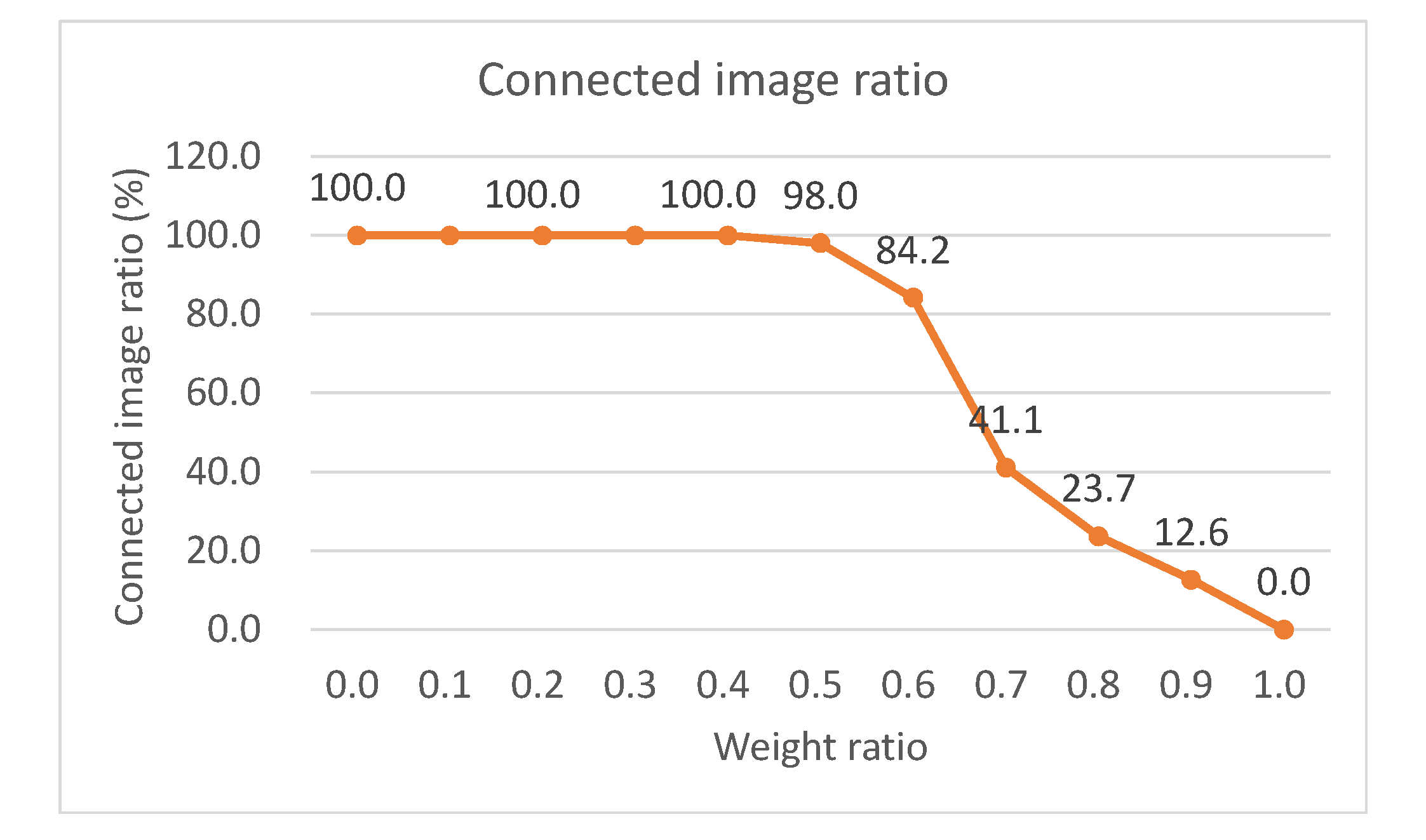}
		\label{fig:figure8-b}}
	\caption{The influence of the weight ratio on global geometric constraint construction for dataset 1.}
	\label{fig:figure8}
\end{figure}

\begin{figure*}[!ht]
	\centering
	\subfloat[Original TCN graph]{\includegraphics[width=0.25\textwidth]{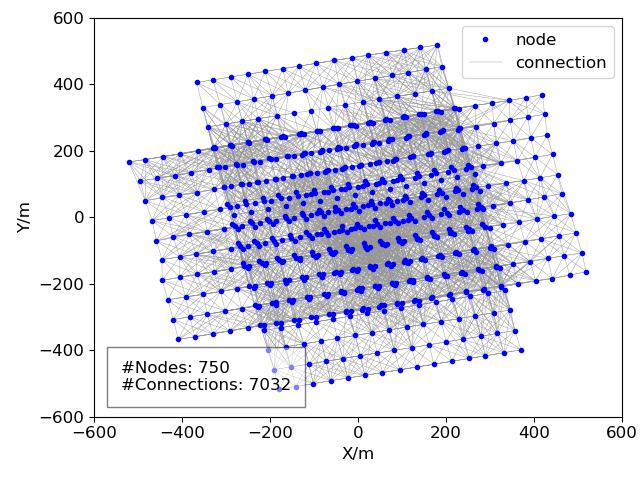}
		\label{fig:figure9-a}}
	\subfloat[WCDS graph with value 0.0]{\includegraphics[width=0.25\textwidth]{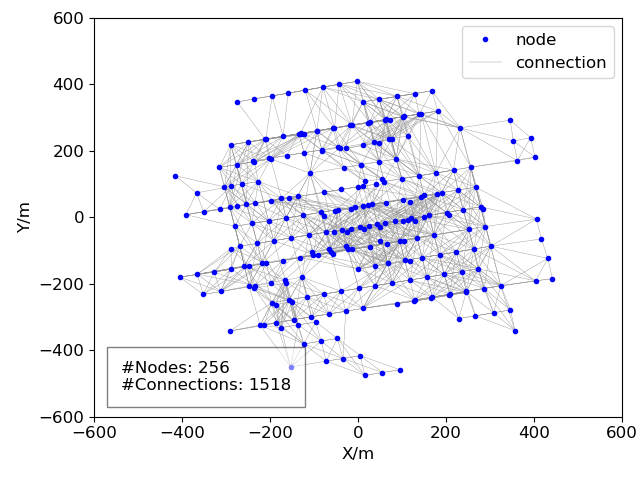}
		\label{fig:figure9-b}}
	\subfloat[WCDS graph with value 0.2]{\includegraphics[width=0.25\textwidth]{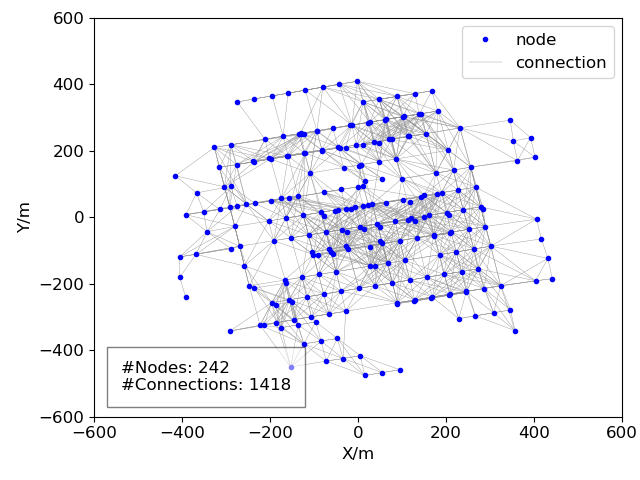}
		\label{fig:figure9-c}} \\
	\subfloat[WCDS graph with value 0.5]{\includegraphics[width=0.25\textwidth]{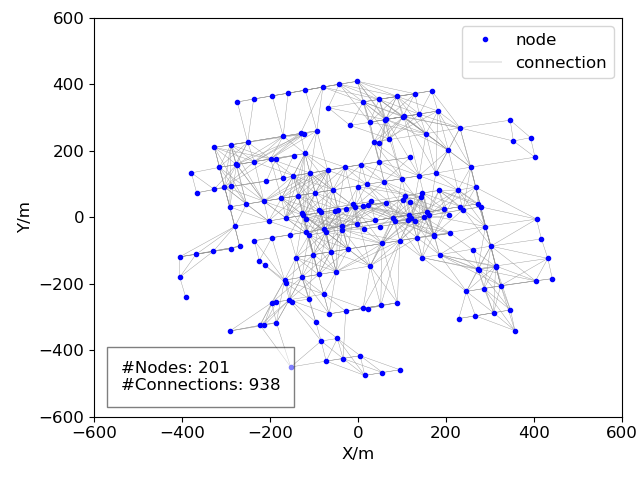}
		\label{fig:figure9-d}}
	\subfloat[WCDS graph with value 0.8]{\includegraphics[width=0.25\textwidth]{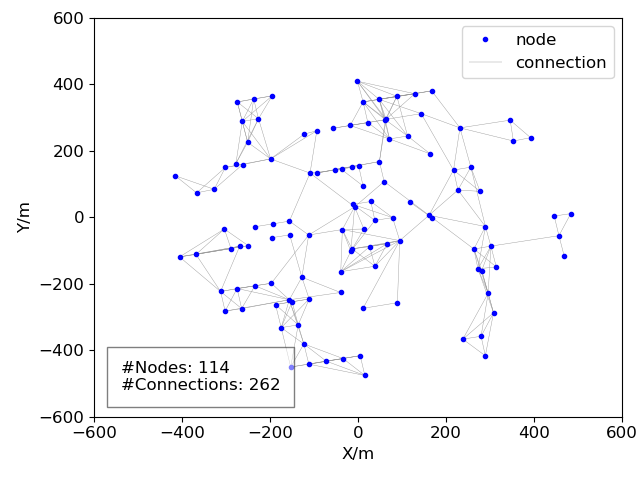}
		\label{fig:figure9-e}}
	\subfloat[WCDS graph with value 1.0]{\includegraphics[width=0.25\textwidth]{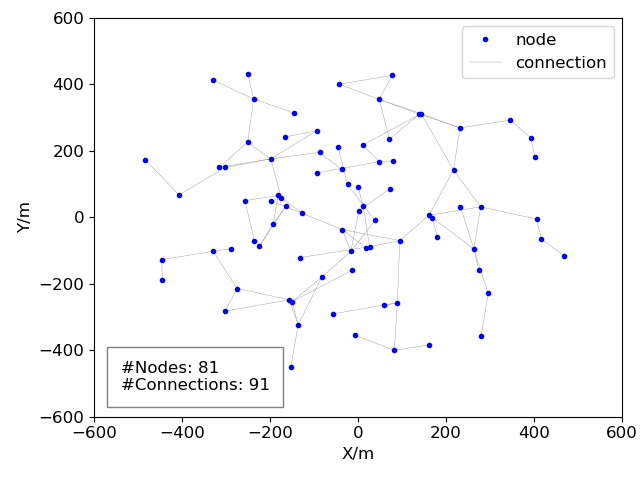}
		\label{fig:figure9-f}}
	\caption{The illustration of the weight ratio on WCDS extraction for dataset 1.}
	\label{fig:figure9}
\end{figure*}

For the analysis of the weight ratio, dataset 1 is selected for experiments, and two metrics are utilized for performance evaluation, i.e., the selected image ratio in WCDS graph and the connected image ratio in SfM reconstruction. The former is the ratio of the number of selected images in the WCDS graph and the number of all images in the dataset; the latter is the ratio of the number of connected images in SfM reconstruction and the number of selected images in the WCDS graph. In this test, the weight ratio is uniformly sampled between 0.0 and 1.0 with the interval value of 0.1. The results are shown in Figure \ref{fig:figure8}, where the influence on the selected image ratio is presented in Figure \ref{fig:figure8-a} and the influence on the connected image ratio is shown in Figure \ref{fig:figure8-b}. It is shown that with the increase of the weight ratio from 0.0 to 1.0, the selected image ratio is almost linearly decreasing from 34.1\% to 10.8\%. The main reason is that by using a smaller weight ratio, WCDS tends to select the next-current vertex that has a stronger edge connection, which further slowdowns the speed of vertex scanning and marking.

By observing the results of the connected image ratio presented in Figure \ref{fig:figure8-b}, we can see that with the increase of the weight ratio from 0.0 to 0.5, almost all selected images have been successfully connected in SfM reconstruction. However, the connected image ratio decreases dramatically when the weight ratio decreases from 0.5 to 1.0. Noticeably, no images have been connected when the weight ratio is set as 1.0. In other words, the extracted images from MCDS has a very weak connection. For visual analysis, Figure \ref{fig:figure9} illustrates the weight ratio on the construction of the WCDS graph, in which Figure \ref{fig:figure9-a} shows the original TCN graph, and Figure \ref{fig:figure9-b} - Figure \ref{fig:figure9-f} show the WCDS graph with the weight ratio of 0.0, 0.2, 0.5, 0.8 and 1.0, respectively. It is clearly shown that the weight ratio has good control on the connection stability of the WCDS graph. By combining the results of the selected image ratio and the connected image ratio, we set the weight ratio $R_{vw}$ as 0.5 in the following tests.

\subsection{The influence of correspondence graph generation on cluster merging}
\label{sec:4.3}

In cluster merging, the correspondence graph is created to build the mapping relationship between feature matches and would be used to find common 3D points across reconstructions. With increasing the number of images, it becomes impossible to build tracks at once for the entire dataset due to the memory limitation and computational costs. In the proposed solution, the on-demand correspondence graph is built by using the feature matches between the images in the source reconstruction and the related images in the reference reconstruction. In this test, we would analyze the influence correspondence graph on cluster merging.

\begin{figure}[t!]
	\centering
	\includegraphics[width=0.45\textwidth]{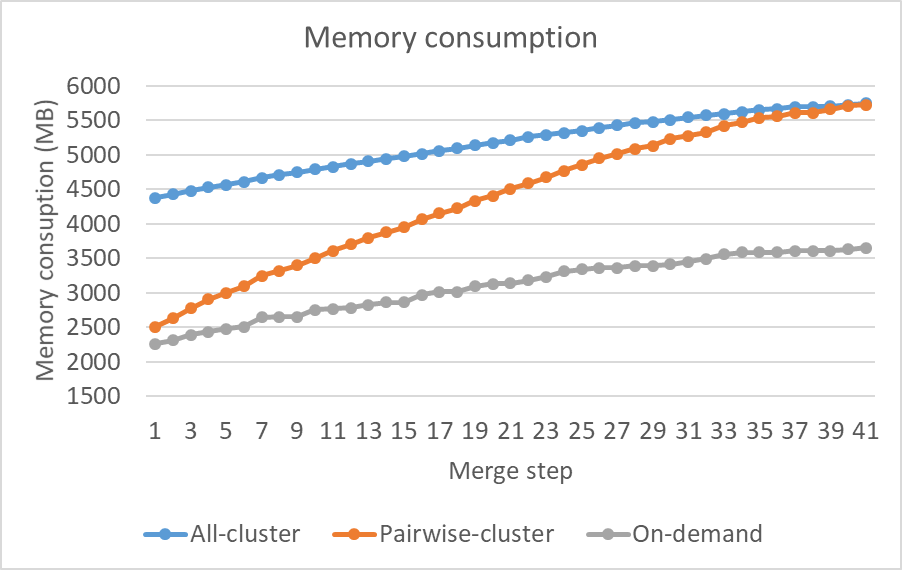}
	\caption{The statistic result of memory consumption during cluster merging for dataset 3.}
	\label{fig:figure10}
\end{figure}

In this section, three solutions for building the correspondence graph have been analyzed, including all-cluster, pairwise-cluster, and on-demand strategies. The first one builds the correspondence graph at once using all feature matches from the dataset; the second one uses all feature matches from two reconstructions; the third one uses feature matches from related images between two reconstructions. For performance evaluation, memory consumption has been used as the metric. Figure \ref{fig:figure10} shows the statistic result in the cluster merging of dataset 3. It is shown that the all-cluster strategy needs the highest memory consumption as it builds the correspondence graph for the entire dataset; although the time consumption of the pair-wise strategy is the same as the proposed on-demand strategy at the beginning, it increases with the number of merged clusters and reaches to the same value as the all-cluster strategy. On the contrary, stable memory consumption can be observed from the on-demand strategy as it only loads required feature matches between reconstructions. For dataset 3, there is approximately 2.1 G memory saving during cluster merging. With the increase of the dataset, more memory saving can be achieved by using the on-demand strategy.

\begin{figure*}[ht!]
	\centering
	\subfloat[]{\includegraphics[width=0.3\textwidth]{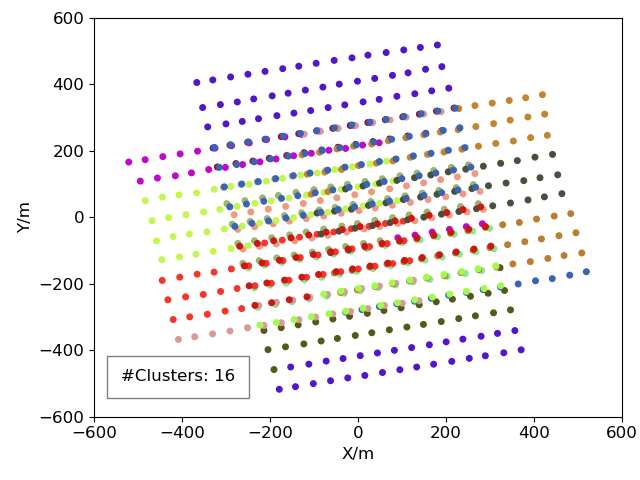}
		\label{fig:figure11-a}}
	\subfloat[]{\includegraphics[width=0.3\textwidth]{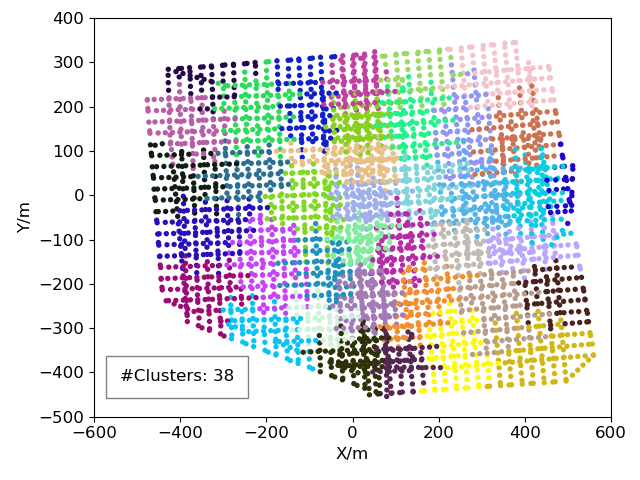}
		\label{fig:figure11-c}}
	\subfloat[]{\includegraphics[width=0.3\textwidth]{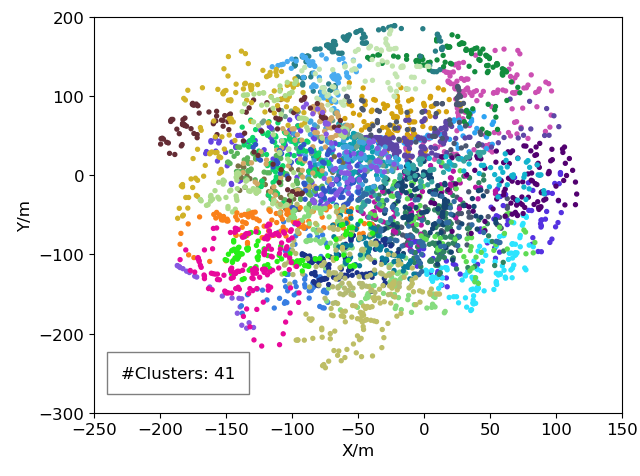}
		\label{fig:figure11-e}} \\
	\subfloat[]{\includegraphics[width=0.3\textwidth]{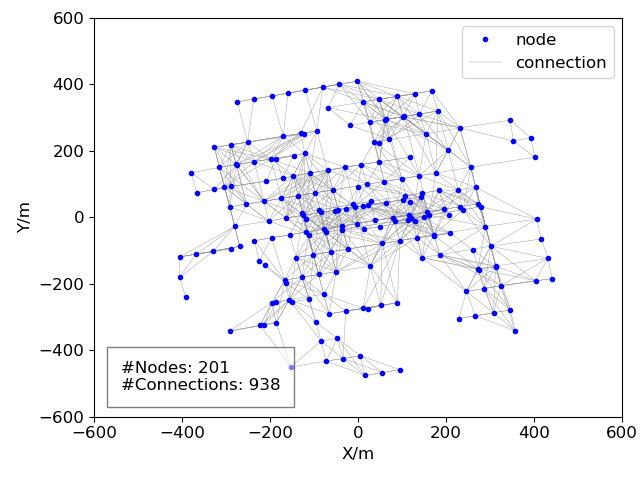}
		\label{fig:figure11-b}}
	\subfloat[]{\includegraphics[width=0.3\textwidth]{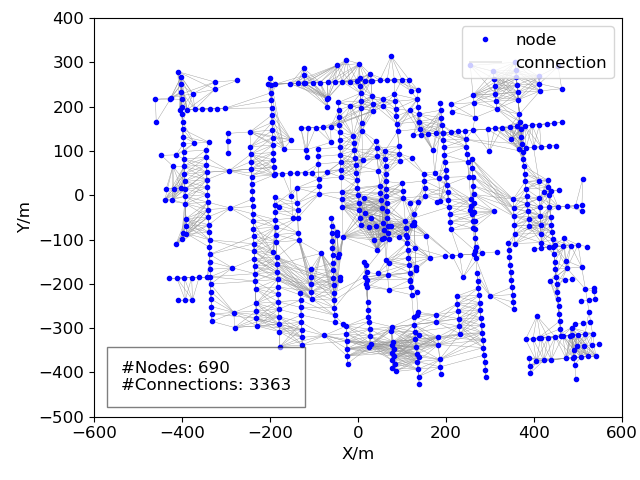}
		\label{fig:figure11-d}}
	\subfloat[]{\includegraphics[width=0.3\textwidth]{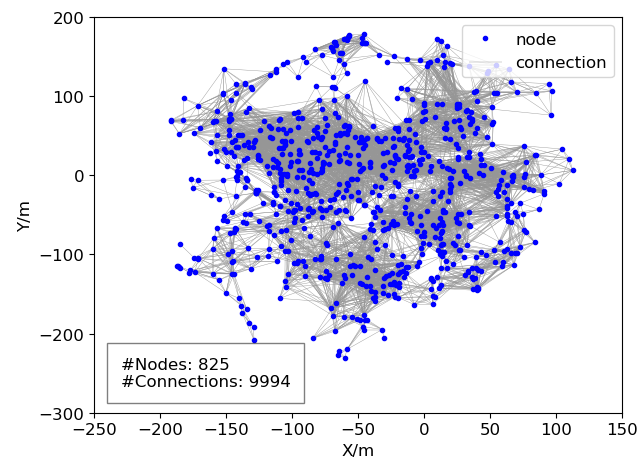}
		\label{fig:figure11-f}}
	\caption{The illustration of scene partition and weighted CDS for datasets 2 and 4: (a), (c), and (e) scene partition with the cluster size of 50, 100, and 100 for these three datasets, respectively; (b), (d) and (f) weighted CDS for the construction of global geometric constraint for the three datasets, respectively.}
	\label{fig:figure11}
\end{figure*}

\subsection{Parallel structure from motion for test datasets}
\label{sec:4.4}

\begin{figure*}[ht!]
	\centering
	\subfloat[197/201]{\includegraphics[height=0.18\textwidth]{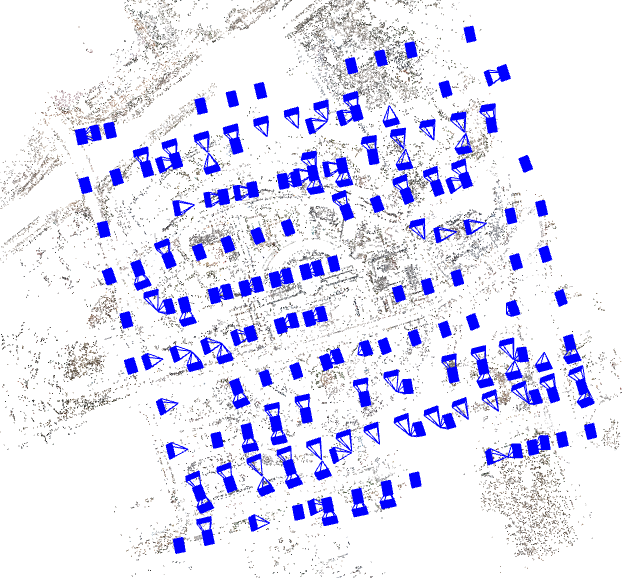}
		\label{fig:figure12-a}}
	\subfloat[750/750]{\includegraphics[height=0.18\textwidth]{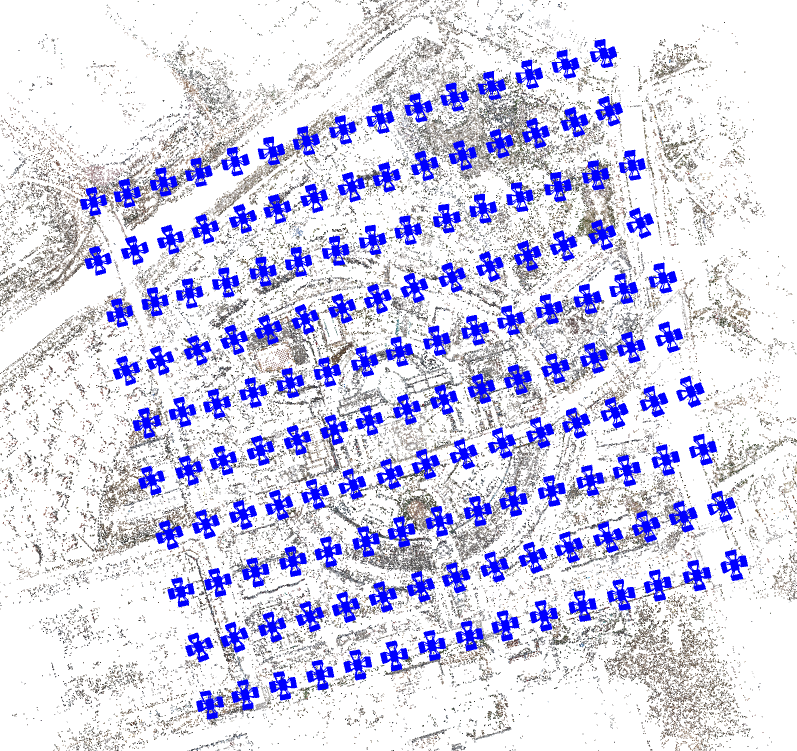}
		\label{fig:figure12-b}}
	\subfloat[684/690]{\includegraphics[height=0.18\textwidth]{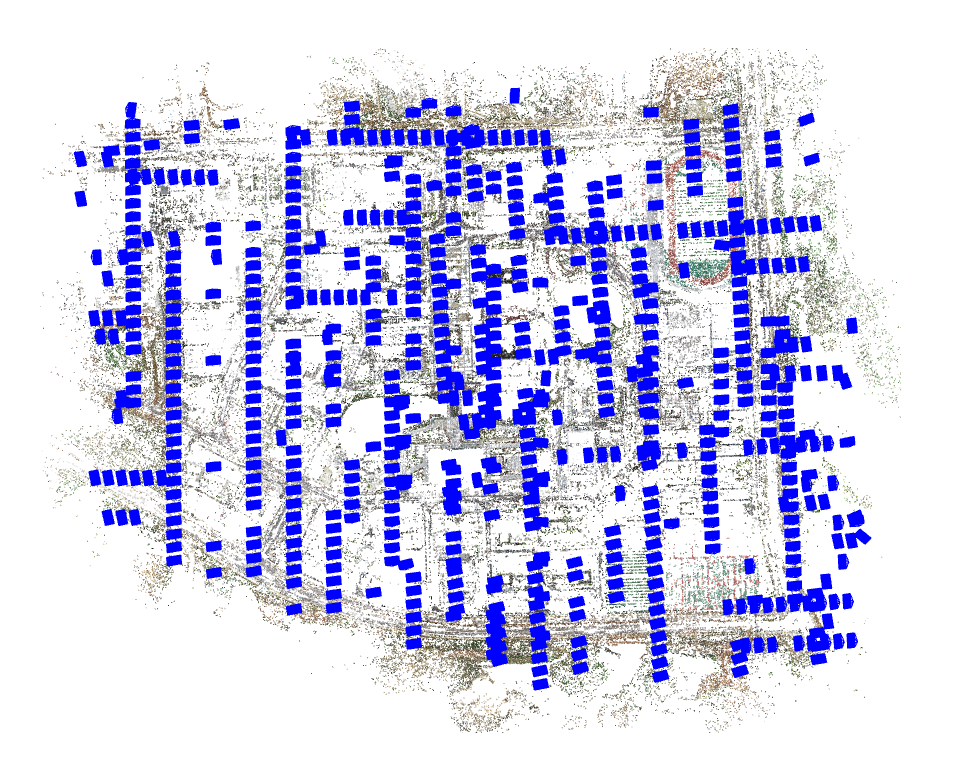}
		\label{fig:figure12-c}}
	\subfloat[3736/3743]{\includegraphics[height=0.18\textwidth]{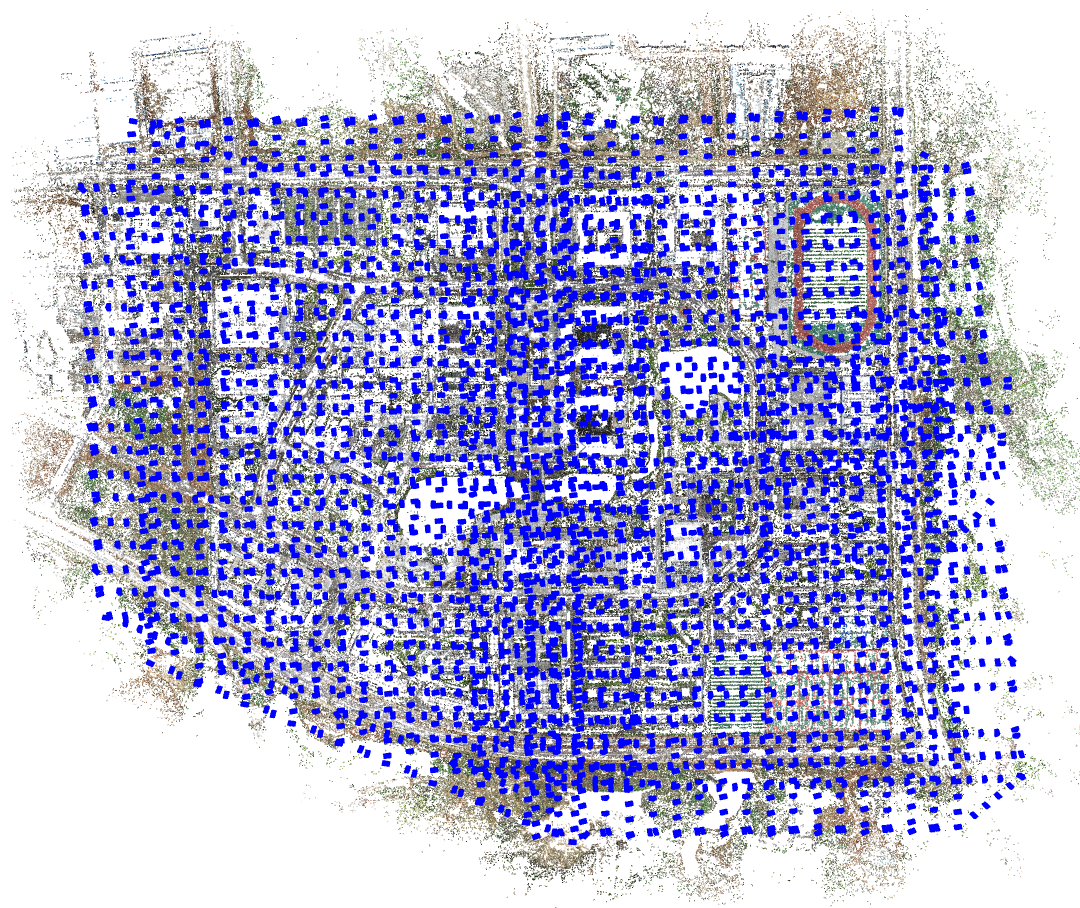}
		\label{fig:figure12-d}} \\
	\subfloat[821/825]{\includegraphics[height=0.2\textwidth]{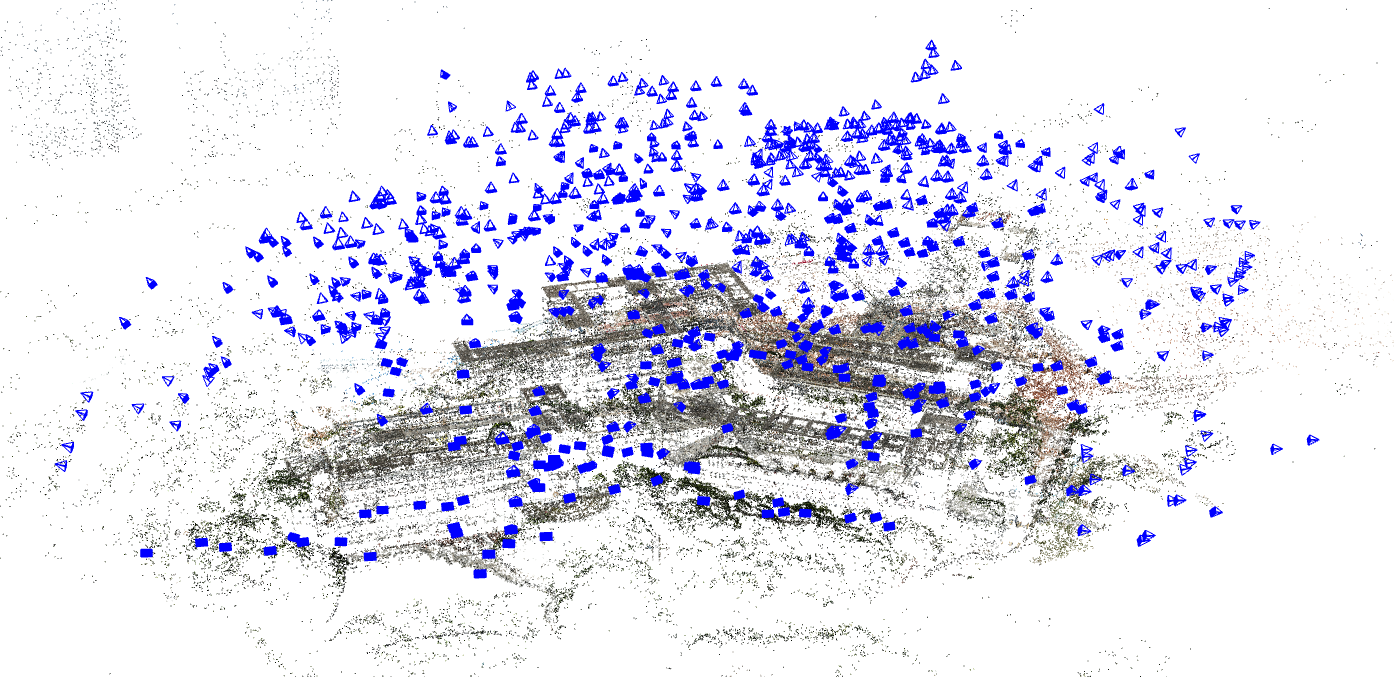}
		\label{fig:figure12-e}}
	\subfloat[3907/4030]{\includegraphics[height=0.2\textwidth]{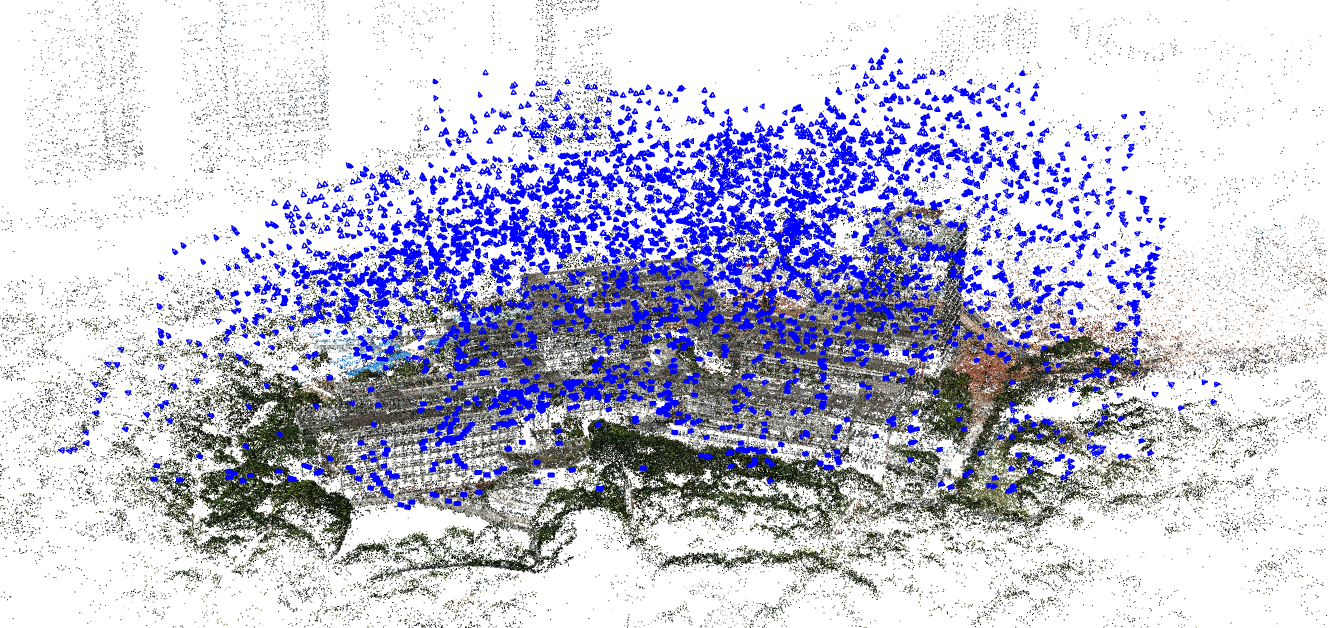}
		\label{fig:figure12-f}}
	\caption{The results of SfM-based 3D reconstruction of the three datasets: (a), (c), and (e) 3D reconstruction results of global geometric models; (b), (d), and (f) 3D reconstruction results of all images in the corresponding datasets. The values in each sub-figure indicate the numbers of resumed and contained images, respectively.}
	\label{fig:figure12}
\end{figure*}

By using the selected optimal parameters, the WCDS graph is extracted to build the global model, are shown in Figure \ref{fig:figure11-b}, Figure \ref{fig:figure11-d}, and Figure \ref{fig:figure11-f}. The number of selected images is 201, 690, and 825 for the three datasets, respectively, which is about 26.8\%, 18.4\%, and 20.5\% of the total number of images in the corresponding dataset. We can also observe that enough image connections have been established in the extracted WCDS graph, which can be verified by the dense gray edges in each WCDS graph. In addition, Figure \ref{fig:figure12-a}, Figure \ref{fig:figure12-c} and Figure \ref{fig:figure12-e} present the SfM reconstructions based on the WCDS graphs, and almost all images in the corresponding graph have been successfully registered, whose number is 197, 684 and 821, respectively.

Meanwhile, scene cluster is also executed to divide the whole scene into clusters. For these three datasets, the number of cluster size is set as 50, 100, and 100, and the number of generated clusters is 16, 38, and 41, which are illustrated in Figure \ref{fig:figure11-a}, Figure \ref{fig:figure11-c} and Figure \ref{fig:figure11-e}, respectively. Noticeably, the cluster result shown in Figure \ref{fig:figure11-e} seems much more irregular than the results presented in Figure \ref{fig:figure11-a}, and Figure \ref{fig:figure11-c}. The main reason is that dataset 3 has been collected by using the optimized views photogrammetry, in which images are not captured at a fixed altitude. After the parallel SfM reconstruction guided by the clusters and the cluster merging constrained by the global model, the final models have been generated and presented in Figure \ref{fig:figure12-b}, Figure \ref{fig:figure12-d}, and Figure \ref{fig:figure12-f} for the three datasets, respectively. We can see that almost all images have been successfully connected for datasets 1 and 2 that are captured by classical oblique photogrammetry. On the contrary, there are 123 images lost in the merged reconstruction of dataset 3, which may be caused by the lacking of enough common 3D points since the irregular data acquisition campaign has been utilized in this dataset.

\subsection{Comparison with state-of-the-art methods}
\label{sec:4.5}

To assess the proposed parallel SfM solution, comparison tests with both open-source and commercial software packages have been conducted in terms of efficiency, precision, and completeness. First, relative BA without GCPs is performed to assess the reconstruction efficiency, relative precision, and completeness of the final model. Second, absolute BA with GCPs is conducted by using ground-truth data to assess the geo-reference accuracy of the final model. For the comparison, two open-source packages, termed ColMap and DagSfM, and two commercial packages, termed Agisoft Metashape and Pix4Dmapper, are used, and these two open-source software packages are implemented by using the C++ programming language. In addition, default parameters are configured for these packages, and all comparison tests are conducted on an Intel Core i7-8700 PC on the Windows platform with 32 GB memory, a 3.19 GHz CPU, and a 6 GB NVIDIAN GeForce GTX 1060 graphics card.

\subsubsection{Relative BA in terms of efficiency, precision, and completeness}
\label{sec:4.5.1}

For relative BA without GCPs, three metrics have been used for performance evaluation, including efficiency, precision, and completeness. The metric efficiency indicates the time costs used in the parallel SfM reconstruction; the metric precision is calculated as the mean reprojection error after the execution of the global BA; the metric completeness is quantified by using the number of registered images and resumed 3D points.

Table \ref{tab:table2} presents the statistical results of relative BA for the three datasets. We can see that the proposed solution achieves the highest efficiency, which is 4.81 min, 100.63 min, and 186.96 min for the three datasets, respectively. The efficiency of DagSfM ranks second since it also utilizes a parallel SfM reconstruction method. When the number of images is not too large, the time costs for the other three packages are acceptable, such as for dataset 1, the time costs are 20.49 min, 43.00 min, and 10.88 min for ColMap, Metashape, and Pix4Dmapper, respectively. However, their efficiency degenerates dramatically with the increase of the involved images. Especially for ColMap, the time cost increases to 3,256.68 min for dataset 3 with 4030 images, which is approximately 17.4 times the time costs consumed by the proposed solution. Even worse, Metashape fails to reconstruct dataset 3 due to the reason being out of memory.

\begin{table*}[t!]
	\centering
	\caption{The statistical results of relative BA without GCPs for the three datasets in terms of efficiency, completeness, and precision. The values in the bracket indicate the number of connected images in the final models.}
	\makebox[\linewidth]{
		\begin{tabular}{lrrrr}
			\toprule
			\textbf{Metric}               & \textbf{Method} & \textbf{Dataset 1} & \textbf{Dataset2} & \textbf{Dataset 3} \\
			\midrule
			\multirow{5}{*}{\begin{tabular}[c]{@{}l@{}}Efficiency\\ (min)\end{tabular}}  & ColMap & 20.49 & 387.84 & 3,256.68 \\
			& DagSfM          & 7.79               & 129.06            & 197.04             \\
			& Metashape       & 43.00              & 208.00            & —                  \\
			& Pix4Dmapper     & 10.88              & 290.72            & 753.90             \\
			& Ours            & 4.81               & 100.63            & 186.96             \\
			\midrule
			\multirow{5}{*}{\begin{tabular}[c]{@{}l@{}}Precision\\ (pixel)\end{tabular}} & ColMap & 0.683 & 0.580  & 0.712    \\
			& DagSfM          & 1.000              & 0.722             & 0.739              \\
			& Metashape       & 0.562              & 0.889             & —                  \\
			& Pix4Dmapper     & 0.253              & 0.379             & 0.373              \\
			& Ours            & 0.410              & 0.374             & 0.429              \\
			\midrule
			\multirow{5}{*}{Completeness} & ColMap          & 308,670 (750)      & 1,211,943 (3,738) & 1,594,456 (4,029)  \\
			& DagSfM          & 252,919 (750)      & 1,258,696 (3,254) & 1,510,435 (3,487)  \\
			& Metashape       & 905,815 (750)      & 7,733,683 (3,742) & —                  \\
			& Pix4Dmapper     & 627,919 (750)      & 4,835,383 (3,743) & 9,987,579 (3,973)  \\
			& Ours            & 231,780 (750)      & 1,253,274 (3,736) & 1,557,104 (3,907) \\
			\bottomrule
		\end{tabular}
		\label{tab:table2}
	}
\end{table*}

\begin{figure*}[htp!]
	\centering
	\subfloat[ColMap]{\includegraphics[height=0.15\textwidth]{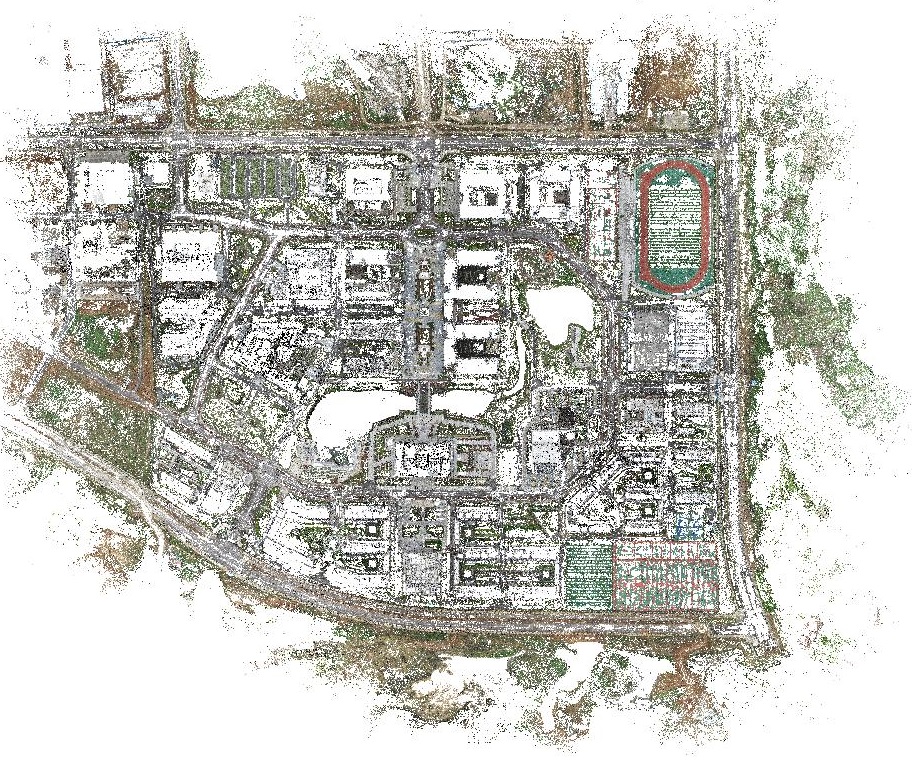}
		\label{fig:figure13-a}}
	\subfloat[DagSfM]{\includegraphics[height=0.15\textwidth]{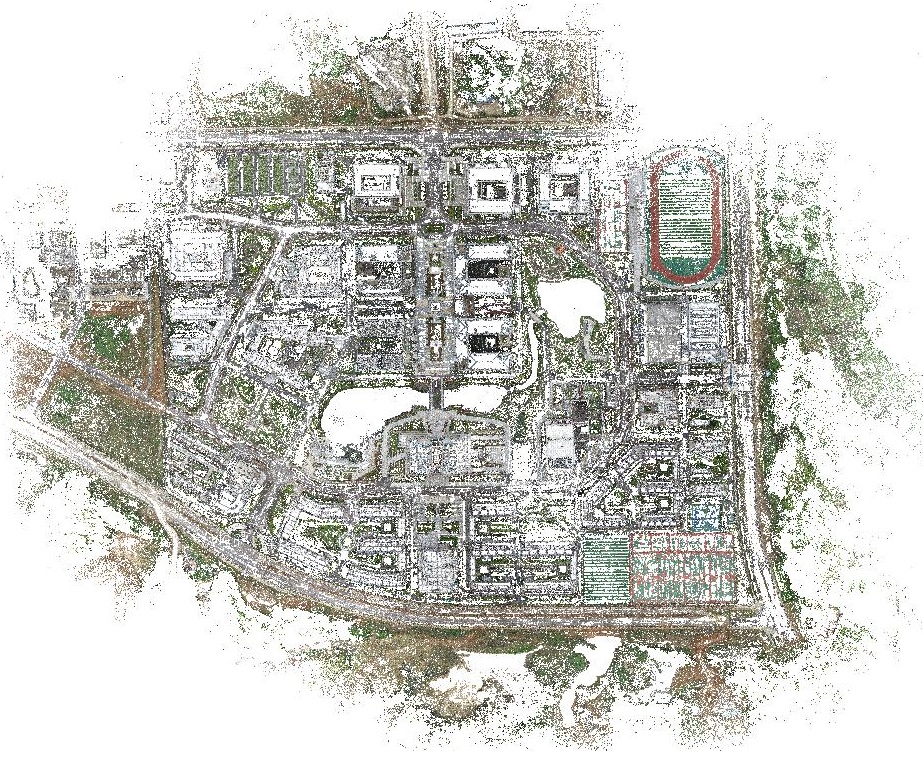}
		\label{fig:figure13-b}}
	\subfloat[Metashape]{\includegraphics[height=0.15\textwidth]{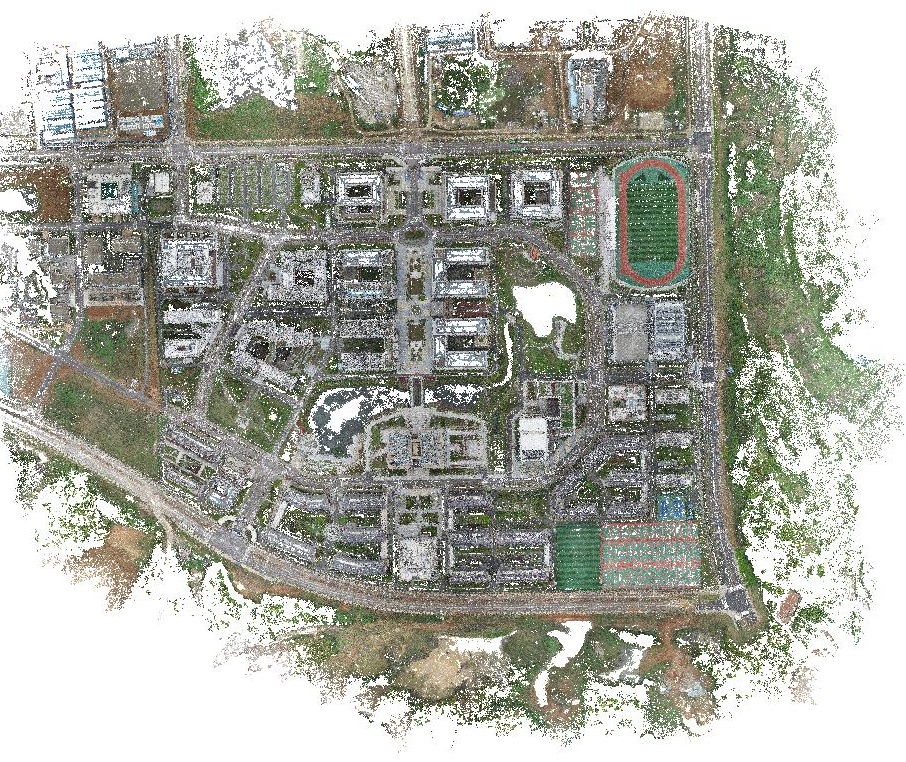}
		\label{fig:figure13-c}}
	\subfloat[Pix4Dmapper]{\includegraphics[height=0.15\textwidth]{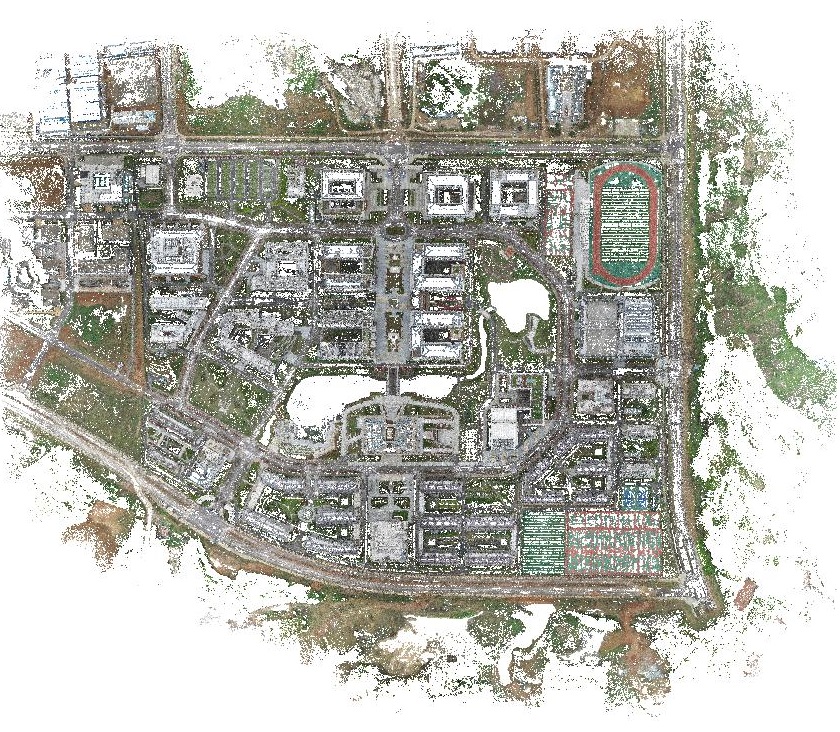}
		\label{fig:figure13-d}}
	\subfloat[Ours]{\includegraphics[height=0.15\textwidth]{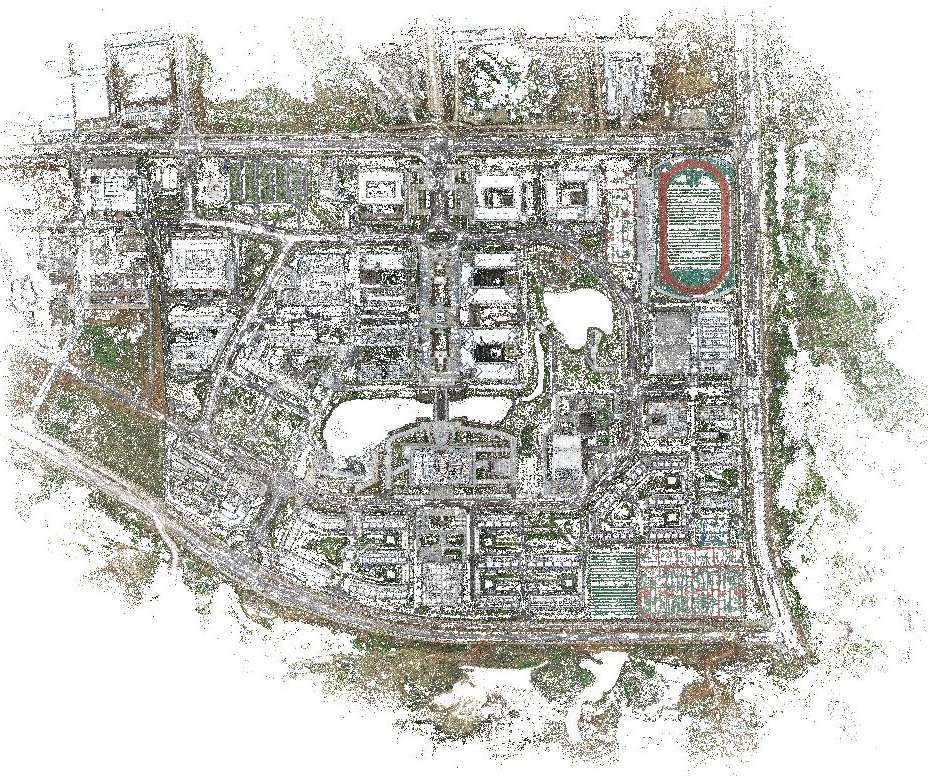}
		\label{fig:figure13-e}}
	\caption{The reconstructed sparse points of dataset 2 based on relative BA without GCPs.}
	\label{fig:figure13}
\end{figure*}

When considering the metrics precision, we can observe that Pix4Dmapper achieves the best performance among all compared packages, which is followed by the proposed solution with the value of 0.410 pixels, 0.374 pixels, and 0.429 pixels for the three datasets. Compared with the parallel method in DagSfM, the proposed solution utilizes the WCDS graph to provide the global geometric constraint in cluster merging and achieves higher BA precision. When considering the metric completeness, it is shown that the largest number of 3D points have been reconstructed from Metashape because it uses the divide-and-conquer strategy for feature detection. For dataset 1 with the smallest image number, all images can be registered by the compared packages. With the increase in the number of involved images, DagSfM lost many more images in the final model when compared with the other packages, which can also be verified by the reconstructed point clouds of dataset 2 as presented in Figure \ref{fig:figure13}. The main reason is that DagSfM merely depends on the overlapped images between reconstructions for cluster merging. By using the global models, the proposed solution registers 3,736 and 3,907 images for datasets 2 and 3, respectively. Based on the results of relative BA, we can conclude that by using the WCDS-based global geometric constraint, the proposed parallel SfM solution can achieve an order of efficiency improvement when compared with classical incremental SfM solution.

\subsubsection{Absolute BA in the term of geo-referencing accuracy}
\label{sec:4.5.2}

With ground-truth data, absolute BA is conducted to evaluate geo-referencing accuracy. During data acquisition of dataset 3, a total number of 26 GCPs have been surveyed and prepared for geo-referencing accuracy assessment. In this test, three GCPs that distribute evenly in the test site are selected to involve the absolute BA for geo-referencing, and all the others are used as check points (CPs). For the accuracy assessment, the residuals of model geo-referencing are calculated as the coordinate differences between CPs and their model points that are triangulated from the geo-referenced model.

\begin{table*}[t!]
	\centering
	\caption{The statistical results of absolute BA with GCPs for dataset 3.}
	\begin{tabular}{llllllllll}
		\toprule
		\multirow{2}{*}{Method} & \multicolumn{3}{l}{Max (m)} & \multicolumn{3}{l}{Mean (m)} & \multicolumn{3}{l}{Std.dev. (m)} \\
		\cline{2-10}
		& $|X|$ & $|Y|$ & $|Z|$ & $|X|$ & $|Y|$ & $|Z|$ & $|X|$ & $|Y|$ & $|Z|$ \\
		\midrule
		ColMap & 0.043 & 0.069 & 0.052 & 0.014 & 0.022 & 0.020 & 0.017 & 0.029 & 0.018 \\
		DagSfM & 0.142 & 0.115 & 0.064 & 0.019 & 0.027 & 0.023 & 0.032 & 0.044 & 0.026 \\
		Pix4dMapper & 0.028 & 0.036 & 0.048 & 0.010 & 0.012 & 0.015 & 0.013 & 0.016 & 0.019 \\
		Ours & 0.040 & 0.052 & 0.080 & 0.016 & 0.020 & 0.036 & 0.020 & 0.024 & 0.031 \\
		\bottomrule
	\end{tabular}
	\label{tab:table3}
\end{table*}

\begin{figure*}[htp!]
	\centering
	\subfloat[]{\includegraphics[width=0.3\textwidth]{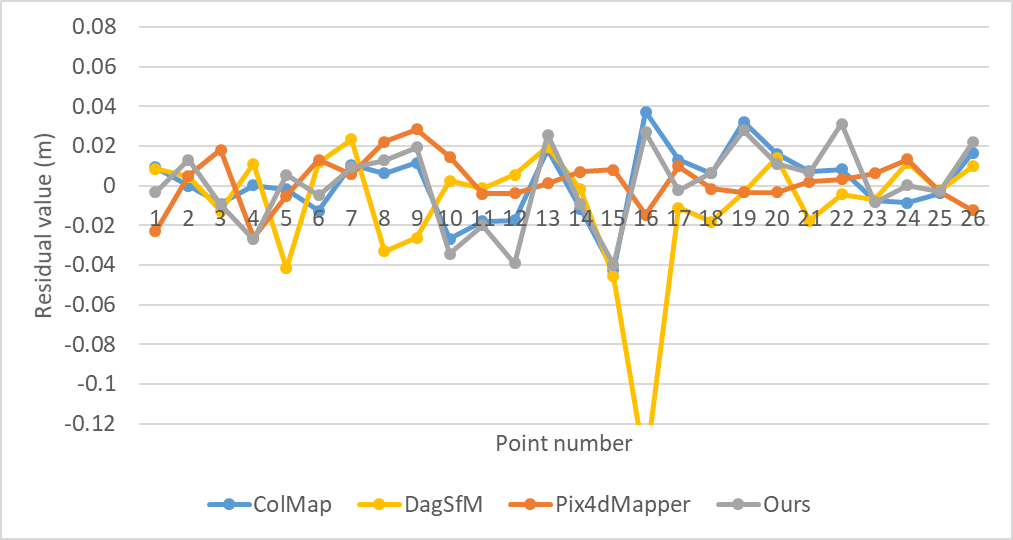}
		\label{fig:figure15-a}}
	\subfloat[]{\includegraphics[width=0.3\textwidth]{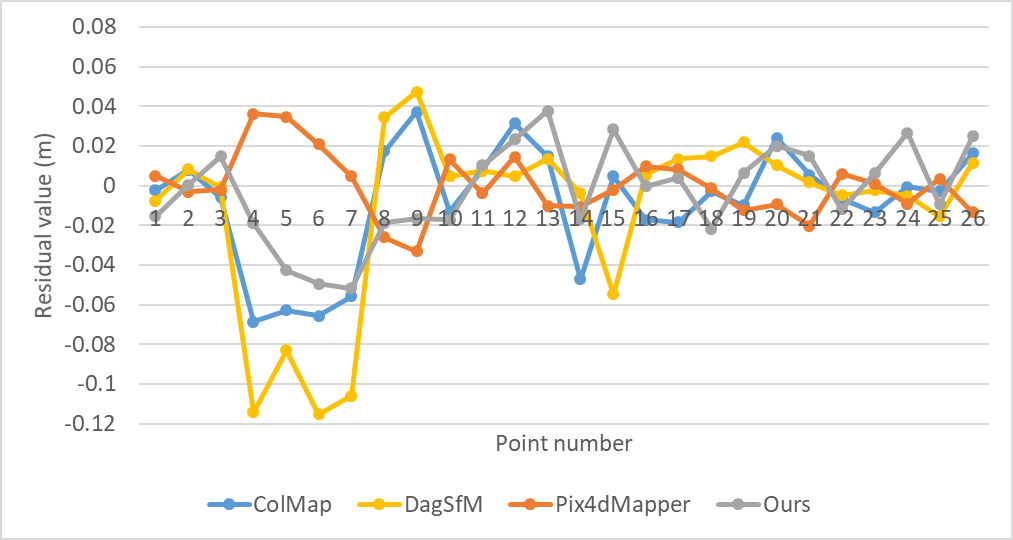}
		\label{fig:figure15-b}}
	\subfloat[]{\includegraphics[width=0.3\textwidth]{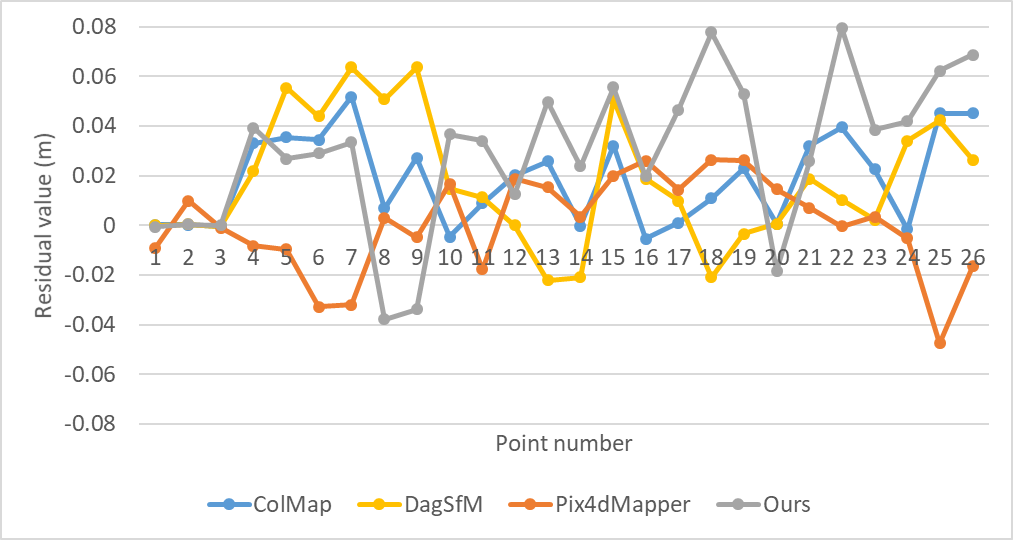}
		\label{fig:figure15-c}}
	\caption{The individual residual plot in the X, Y, and Z directions for dataset 3: (a) the residual plot in the X direction; (b) the residual plot in the Y direction; (c) the residual plot in the Z direction.}
	\label{fig:figure15}
\end{figure*}

Table \ref{tab:table3} shows the statistical results of geo-referencing accuracy, and three metrics, termed max, mean and std.dev. of absolute residuals, are used for performance comparison. We can see that Pix4Dmapper achieves the highest accuracy among all compared methods, whose residual in the term of std.dev. is 0.013 m, 0.016 m, and 0.019 m in the X, Y, and Z directions, respectively. Although the vertical accuracy of ColMap ranks second with the std.dev. value of 0.018 m, the proposed solution can achieve comparative accuracy in the horizontal directions with the std.dev. value of 0.020 m, and 0.024 m in the X and Y directions, respectively, which can also be demonstrated by the individual residual plots as presented in Figure \ref{fig:figure15-a} and Figure \ref{fig:figure15-b}. In addition, when compared with DagSfM, the proposed solution has achieved better accuracy in the horizontal direction for the three metrics. Considering that the GSD of dataset 3 is 1.2 cm, the overall geo-referencing accuracy of the proposed solution is approximately 1.7, 2.0, and 2.6 times the GSD value in the X, Y, and Z directions. Based on the results of absolute BA, we can conclude that by using the WCDS-based global geometric constraint, the proposed parallel SfM solution can achieve more evenly distributed and comparative geo-referencing accuracy in the horizontal direction.

\section{Conclusions}
\label{sec:5}

In this paper, we have proposed the WCDS algorithm for the global model extraction and a parallel SfM solution to implement efficient and accurate 3D reconstruction for UAV images. The core idea of the proposed solution is to create a global model to facilitate cluster merging. The experimental results from real UAV images demonstrate that the proposed parallel SfM can dramatically increase the reconstruction efficiency when compared with both open-source and commercial software packages, and comparative orientation accuracy has also been achieved in both relative and absolute BA tests.

Some limitations and possible improvements have also been observed in this study. First, the proposed SfM solution depends on the global model to assist scene merging. In the experiments as presented in Section \ref{sec:4.2}, there are approximately 20\% of images retained to build the global model. With the increase of datasets, the volume of the global model would become too large and then degenerates the overall performance of the parallel solution. To overcome the limitation, the global model would cooperate with the hierarchical strategy \cite{toldo2015hierarchical} to avoid increasing growth. Second, by the further analysis of time consumption, we found that approximately half of the time costs are consumed in the scene merging and the final BA optimization due to two main reasons. On the one hand, the bi-directional mean square reprojection error has been used in the RANSAC framework for the similarity transformation estimation, which is more time consuming than the direct point cloud-based methods; on the other hand, all camera poses and resumed 3D points participate in the final optimization, which causes high computation. To cope with this situation, future work would focus on designing an efficient strategy for similarity transformation estimation and selecting the most valuable tracks for the final BA optimization \cite{cao2019fast, zhang2016efficient}.

\section*{Acknowledgment}
\label{Acknowledgment}

The authors would like to thank the anonymous reviewers and editors, whose comments and advice improve the quality of the work. This research was funded by the National Natural Science Foundation of China (Grant No. 42001413).



\ifCLASSOPTIONcaptionsoff
  \newpage
\fi



%
%
%

\bibliographystyle{IEEEtran}
\nocite{*}
\bibliography{mybibfile}

%

\begin{IEEEbiography}[{\includegraphics[width=1in,height=1.25in,clip,keepaspectratio]{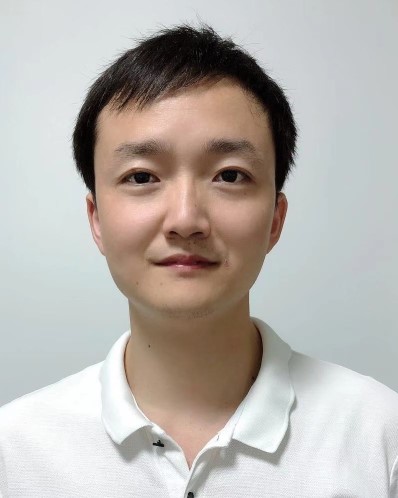}}]{San Jiang}
received the B.S. degree in remote sensing science and technology from Wuhan University in 2010, and the M.Sc. and Ph.D. degrees in photogrammetry and remote sensing from Wuhan Univeristy in 2012 and 2018, respectively. From 2012 to 2014, he worked as an assistant engineer in Tianjin Institute of Surveying and Mapping. From 2014 to 2015, he joined the LIESMARS (State Key Laboratory of Information Engineering in Surveying, Mapping and Remote Sensing of Wuhan Univeristy) as a research assistant.

Currently, he is an associate professor in the School of Computer Science at China University of GeoSciences (Wuhan). His research interests include image matching, SfM-based aerial triangulation, and 3D reconstruction.
\end{IEEEbiography}

\begin{IEEEbiography}[{\includegraphics[width=1in,height=1.25in,clip,keepaspectratio]{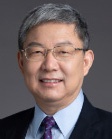}}]{Qingquan Li}
received the Ph.D. degree in Geographic
Information System (GIS) and photogrammetry
from the Wuhan Technical University of Surveying and Mapping, Wuhan, China, in 1998.
From 1988 to 1996, he was an Assistant Professor with Wuhan University, Wuhan, where he became an Associate Professor, in 1996, and has been a Professor, since 1998. He is currently a Professor of Shenzhen University, Shenzhen, China; a Professor with the State Key Laboratory of Information Engineering in Surveying, Mapping and Remote Sensing, Wuhan University; and the Director of Shenzhen Key Laboratory of Spatial Smart Sensing and Service, Shenzhen. His research interests include intelligent transportation systems, 3-D and dynamic data modeling, and pattern
recognition.

Dr. Li is an Academician of International Academy of Sciences for Europe
and Asia (IASEA), an Expert in Modern Traffic with the National 863 Plan,
and an Editorial Board Member of the Surveying and Mapping Journal and
the Wuhan University Journal—Information Science Edition.
\end{IEEEbiography}

\begin{IEEEbiography}[{\includegraphics[width=1in,height=1.25in,clip,keepaspectratio]{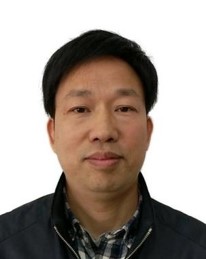}}]{Wanshou Jiang}
received his bachelor and master degrees in photogrammetry and remote sensing from Wuhan Technical University of Surveying and Mapping respectively in 1989 and 1996. In 2004, he received the PhD degree in photogrammetry and remote sensing from Wuhan University. He started his research career in 1989 as a software developer in analytical photogrammetry. In 2000, he joined the LIESMARS (the State Key Laboratory of Information Engineering in Surveying, Mapping and Remote Sensing) as an associate researcher and then he got the tenure position of researcher in 2005.
	
His research interest includes image registering, image classification, change detection, 3D reconstruction, etc. He made a lot of contribution to the famous digital photogrammetric workstation VirtuoZo and designed a software platform, named OpenRS, for remote sensing image processing.
\end{IEEEbiography}

\begin{IEEEbiography}[{\includegraphics[width=1in,height=1.25in,clip,keepaspectratio]{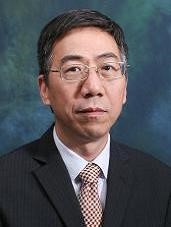}}]{Wu Chen}
 a Professor with the Department of Land
Surveying and Geo-Informatics of the Hong Kong
Polytechnic University. He has been actively working on GNSS related research for more than 30 years. His main research interests are geodesy and geodynamics, seamless positioning technologies, GNSS positioning and applications, system integration, GNSS performance evaluation, regional GPS network, and SLAM.
\end{IEEEbiography}





\end{document}